\documentclass[10pt,twocolumn,letterpaper]{article}

\usepackage{iccv}
\usepackage{times}
\usepackage{epsfig}
\usepackage{graphicx}
\usepackage{amsmath}
\usepackage{amssymb}
\usepackage{subcaption}
\usepackage{pifont}
\usepackage{tabu,multirow}
\usepackage{booktabs}
\usepackage[pagebackref=true,breaklinks=true,letterpaper=true,colorlinks,bookmarks=false]{hyperref}
\usepackage{pdfpages}

\usepackage[table]{xcolor}
\definecolor{LightCyan}{rgb}{0.88,1,1}
\newcommand{\cmark}{\ding{51}}%
\newcommand{\xmark}{\ding{55}}%

\newcommand{\cityscapes}{Cityscapes}
\newcommand{\kitti}{Kitti}
\newcommand{\synthia}{Synthia}
\newcommand{\carla}{Carla}

\newcommand{\semdep}{$Sem. \rightarrow Dep.$}
\newcommand{\depsem}{$Dep. \rightarrow Sem.$}

\newcommand{\domainA}{$\mathcal{A}$}
\newcommand{\domainB}{$\mathcal{B}$}
\newcommand{\trnet}{$G_{1\rightarrow2}$}
\newcommand{\sourcenet}{$N_1^{A\cup{B}}$}
\newcommand{\targetnet}{$N_2^A$}
\newcommand{\taskone}{$\mathcal{T}_1$}
\newcommand{\tasktwo}{$\mathcal{T}_2$}

\newcommand{\algoname}{AT/DT}
\newcommand{\algonamelong}{Across Task and Domain Transfer Framework}

\iccvfinalcopy % *** Uncomment this line for the final submission

\def\iccvPaperID{3494} % *** Enter the ICCV Paper ID here

\ificcvfinal\fi
\begin{document}
	
	%%%%%%%%% TITLE
	\title{Learning Across Tasks and Domains}  % and Domain??%
	
	\author{Pierluigi Zama Ramirez , Alessio Tonioni, Samuele Salti, Luigi Di Stefano\\
		Department of Computer Science and Engineering (DISI)\\
		University of Bologna, Italy\\
		{\tt\small \{pierluigi.zama, alessio.tonioni, samuele.salti, luigi.distefano\}@unibo.it}
	}

	\maketitle
	%\thispagestyle{empty}

	%%%%%%%%% ABSTRACT
	\begin{abstract}
		Recent works have proven that many relevant visual tasks are closely related one to another.
		Yet,  this connection is seldom deployed in practice due to the lack of practical methodologies to transfer learned concepts across different training processes. In this work, we introduce a novel adaptation framework that can operate across both task and domains.
		Our framework learns to transfer knowledge across tasks in a fully supervised domain (\eg{,} synthetic data) and use this knowledge on a different domain where we have only partial supervision (\eg{,} real data). 
		Our proposal is complementary to existing domain adaptation techniques and extends them to cross tasks scenarios providing additional performance gains. 
		We prove the effectiveness of our framework across two challenging tasks (\ie{,} monocular depth estimation and semantic segmentation) and four different domains (\synthia{}, \carla{}, \kitti{}, and \cityscapes{}).    
	\end{abstract}
	
	%%%%%%%%% BODY TEXT
	\section{Introduction}
	
	Deep learning has revolutionized computer vision research and set forth a general framework to address a variety of visual tasks (\eg, classification, depth estimation, semantic segmentation, \dots). 
	The existence of a common framework suggests a close relationship between different tasks that should be exploitable to alleviate the dependence on huge labeled training sets.
	Unfortunately, most state-of-the-art methods ignore these connections and instead focus on a single task by solving it in isolation through supervised learning on a specific domain (\ie, dataset).
	Should the domain or task change, common practice would require acquisition of a new annotated training set followed by retraining or fine-tuning the model. 
	%However, we all know that \emph{it's a long way to the top to label datasets}.
	However, any deep learning practitioner can testify that the effort to annotate a dataset is usually quite substantial and does vary significantly across tasks, potentially requiring ad-hoc acquisition modalities. 
	The question we try to answer is: \emph{would it be possible to deploy the relationships between tasks to remove the dependence for labeled data on new domains?}
	
	\begin{figure}
		\centering
		\includegraphics[width=0.47\textwidth]{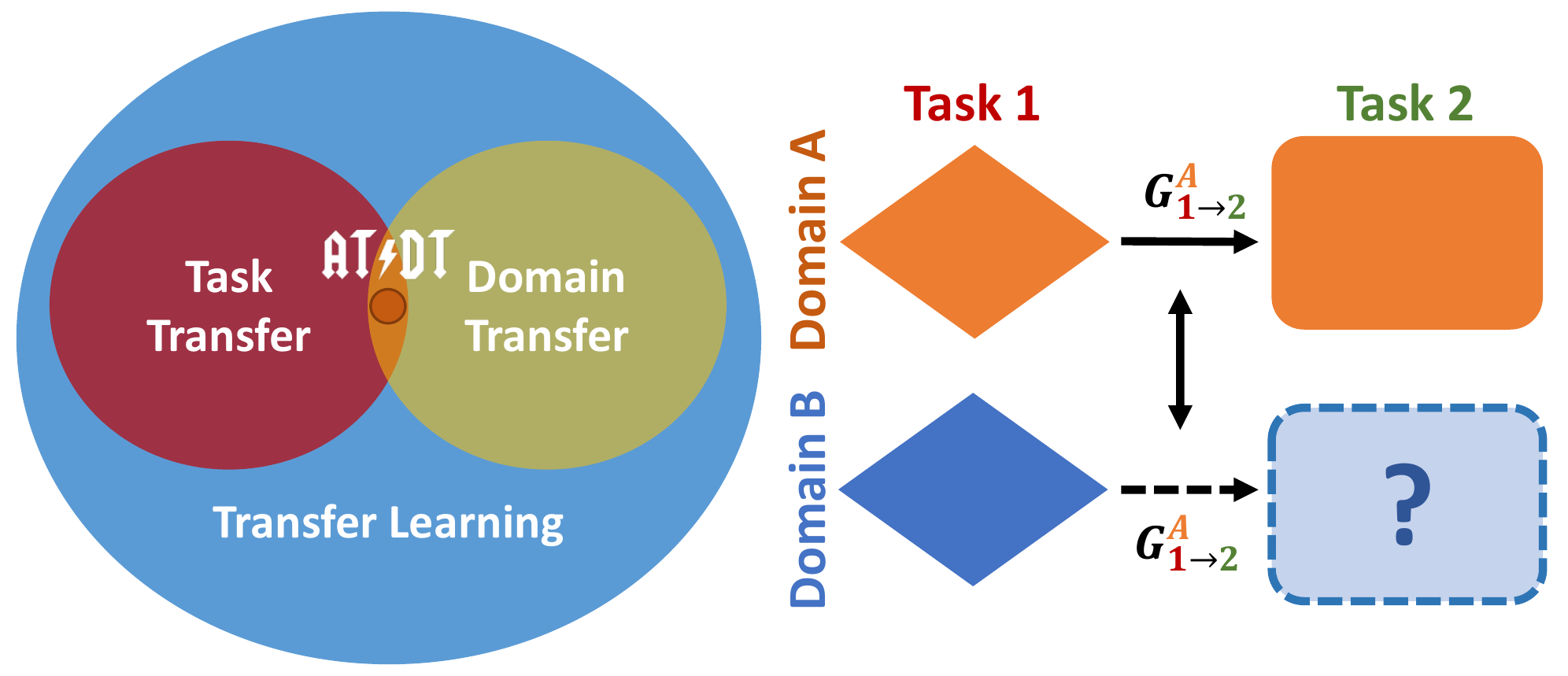}
		\caption{Our \algoname{} framework transfers knowledge across tasks and domains. Given two tasks (1 and 2) and two domains (A and B), with supervision for both tasks in A but only for one task in B, we learn the dependency between tasks in A and exploit this knowledge in B to solve task 2 without the need of supervision.}
		\label{fig:teaser}
	\end{figure}
	
	A partial answer to this question has been provided by \cite{zamir2018taskonomy}, which formalizes the relationships between tasks within a specific domain into a graph referred to as \emph{Taskonomy}. 
	This knowledge can be used to improve performance within a fully supervised learning scenario, though it is not clear how well may it generalize to new domains and to which extent may it be deployed in a partially supervised scenario (\ie, supervision on only some tasks/domains). 
	Generalization to new domains is addressed in the domain adaptation literature \cite{wang2018deep}, that, however, works under the assumption of solving a single task in isolation, therefore ignoring potential benefits from related tasks.
	
	We fuse the two worlds by explicitly addressing a cross domain and cross task problem where on one domain (\eg{,} synthetic data) we have annotations for many tasks, while in the other (\eg{,} real data) annotations are available only for a specific task, though we wish to solve many.
	
	Purposely, we propose a new `\underline{A}cross \underline{T}asks and \underline{D}omains \underline{T}ransfer framework' (shortened \algoname{}	\footnote{Code for \algoname{} released at  \small{\url{https://github.com/CVLAB-Unibo/ATDT}.}}) which learns in a specific domain a function \trnet{} to transfer knowledge between a pair of tasks.
	After the training phase, we show that the same function can be applied in a new domain to solve the second task while relying on supervision only for the first.
	A schematic representation of \algoname{} is pictured in \autoref{fig:teaser}.
	
	We prove the effectiveness of \algoname{} on a challenging autonomous driving scenario where we address the two related tasks of depth estimation and semantic segmentation \cite{ramirez2018geometry}. 
	Our framework allows the use of fully supervised synthetic datasets (\ie, Synthia \cite{HernandezBMVC17}, and Carla \cite{Dosovitskiy17}) to drastically boost performance on partially supervised real data (\ie, Cityscapes \cite{Cordts_2016_CVPR} and \kitti{} \cite{KITTI_2012,KITTI_2015}).
	Finally, we will also show how \algoname{} is robust to sub-optimal scenarios where we use only few annotated real samples or noisy supervision by proxy labels \cite{klodt2018supervising,yang2018deep,tonioni2017unsupervised}.
	The novel contributions of this paper can be summarized as follows:
	\begin{itemize}
		\itemsep0em
		\item According to the definition of task in \cite{zamir2018taskonomy}, to the best of our knowledge we are the first to study a cross domain and cross task problem where supervision for all tasks is available in one domain whilst only for a subset of them in the other.
		%\item To the best of our knowledge, we are the first to study a cross task and cross domain problem where we have supervision for all tasks in one domain while only for a subset of them in the other. 
		\item  We propose a general framework to address the aforementioned problem that learns to transfer knowledge across tasks and domains
		\item We show that it is possible to directly learn a mapping between features suitable for different tasks in a first domain and that this function generalizes to unseen data both in the source domain as well as in a second target domain.
		%\item We show that it is possible to directly learn a mapping between representations suitable for different tasks and that this function generalizes well to unseen data both in the same domain as well as in new ones.
	\end{itemize}
	
	\section{Related Work}
	%We subdivide related works into three paragraphs.
	\textbf{Transfer Learning:} 
	The existence of related representation within CNNs trained for different tasks has been highlighted since early works in the field   \cite{yosinski2014transferable}. 
	These early findings have motivated the use of transfer learning strategy to bootstrap learning across related tasks. For example,  object detection networks are typically initialized with Imagenet weights \cite{Ren_2017, He_2017, liu2016ssd}, although  \cite{he2018rethinking} has recently challenged this paradigm.  
	Luo et al. \cite{luo2017label} fuse transfer learning  with domain adaptation to transfer representations across tasks and domains. However, their definition of tasks deals with different sets of classes in a classification problem, while we consider diverse visual tasks as set forth in \cite{zamir2018taskonomy}.
	Recently Zamir et. al. \cite{zamir2018taskonomy} have tried to formalize and deploy the idea of reusing information across training processes by proposing a computational approach to establish relationships among visual tasks represented in a taxonomy. 
	Pal et. al. \cite{pal2019zeroshot}, propose to use  similar knowledge alongside with meta-learning to learn how to perform a new task within a zero-shot scenario.
	Both \cite{zamir2018taskonomy} and \cite{pal2019zeroshot} assume a shared domain across the addressed task,  whilst  we directly target a cross domain scenario. 
	Moreover, \cite{zamir2018taskonomy} assumes full supervision to be available for all tasks while \cite{pal2019zeroshot} zero supervision for the target task, 
	Differently, our work leverages on full supervision for all tasks in one domain and only partial supervision in a different (target) domain.
	
	\textbf{Multi-task Learning:}
	Multi-task learning tries to learn many tasks simultaneously to obtain more general models or multiple outputs in a single run \cite{Kokkinos_2017, Doersch_2017, Guo_2018_ECCV}. 
	Some recent works have addressed the autonomous driving scenario \cite{chennupati2019auxnet,ramirez2018geometry} so to learn jointly related tasks like depth estimation and semantic segmentation in order to improve performance. 
	Kendall et al. \cite{Cipolla_2018} additionally consider the instance segmentation task and show that it is possible to train a single network to solve the three tasks by fusing the different losses through uncertainty estimation. 
	Our work, instead, directly targets a single task but tries to use the relationship between related tasks to alleviate the need for annotations.
	
	\textbf{Domain Adaptation:}
	A recent survey of the domain adaptation literature can be found in \cite{Wang_2018}. The idea behind this field is to learn models that turn out robust when tested on data sampled from a domain different than the training one.
	Earliest approaches such as \cite{gong2012geodesic,gopalan2011domain} try to build intermediate representations across domains, while recent ones, specifically designed for deep learning, focus on adversarial training at either pixel or feature level. 
	Pixel level methods \cite{shrivastava2017learning,Zheng_2018_ECCV,Bousmalis_2017, ramirez2018exploiting} seek to transform input images from one domain into the other one using recently proposed image-to-image translation GANs \cite{Zhu_2017_ICCV,Isola_2017_CVPR}.
	Conversely, feature level methods  \cite{hoffman2016fcns,long15Learning,tzeng2017adversarial,tzeng2015simultaneous,ganin2015unsupervised,zhang2017curriculum, Tsai_2018} try to align the feature representation extracted from CNN across different datasets, usually, again, by using GANs. 
	Finally, recent works \cite{Sankaranarayanan_2018,hoffman2017cycada, Zhang_2018} operate at both pixel and feature level
	and focus on a single specific task (usually semantic segmentation), while our framework leverages  information from different tasks.
	As such, we argue that our new formulation can be seen as complementary to existing domain adaptation techniques.  
	
	\section{\algonamelong{}}
	
	\begin{figure*}[t]
	\centering
	\includegraphics[width=0.9\textwidth]{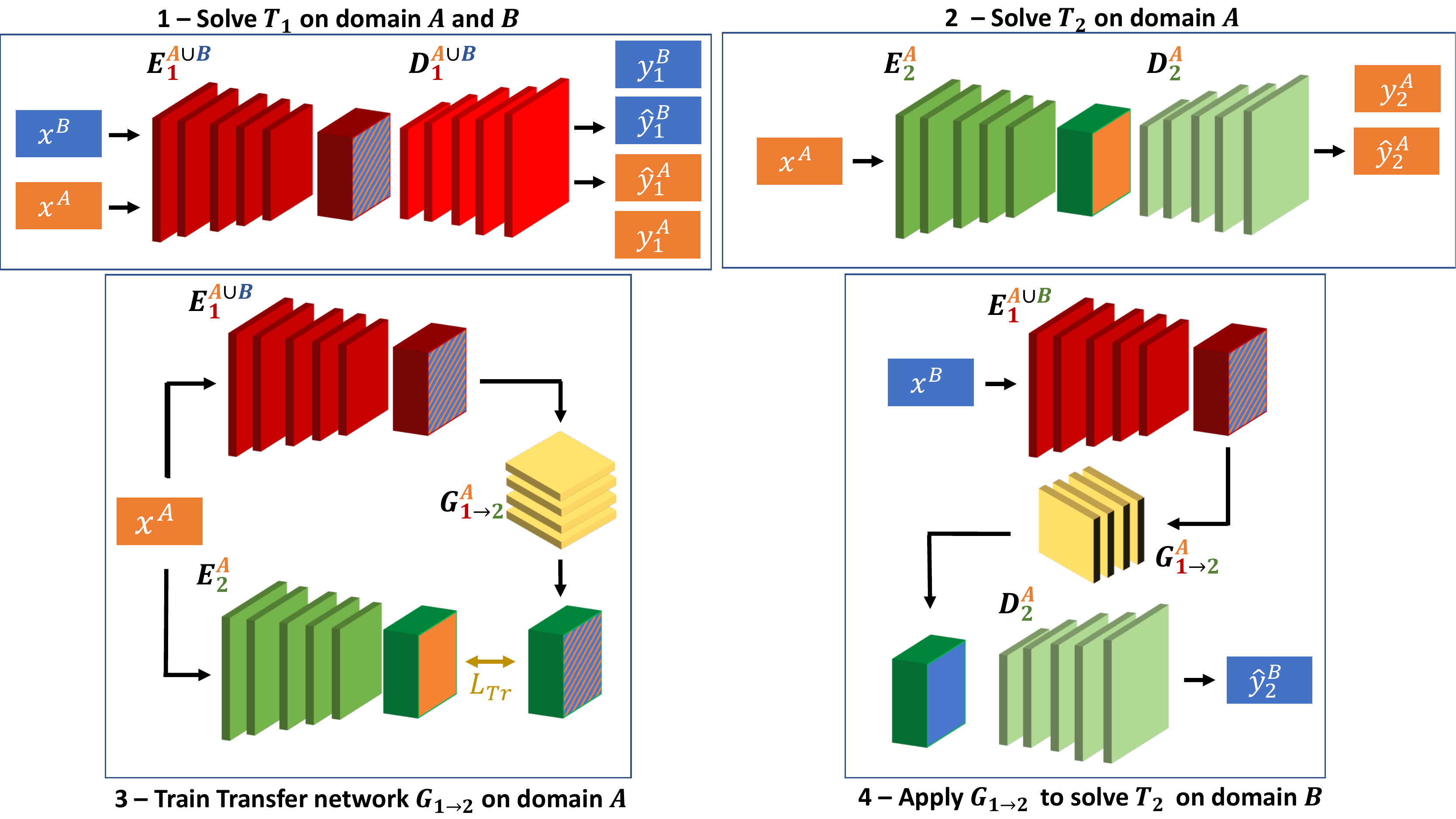}
	\caption{Overview of the \algoname{} framework.
	% We propose a four steps framework which allows to transfer knowledge across tasks and domains.
	(1) We train network \sourcenet{} to solve \taskone{} (red) with supervision in domain \domainA{} (orange) and \domainB{} (blue) to obtain a shared feature representation across domains, highlighted by blue and orange strips. (2) We train a network \targetnet{} to solve \tasktwo{} (green) on \domainA{} where labels are available. (3) We learn a network \trnet{} that transform features from  \taskone{} to  \tasktwo{} on samples from \domainA{}. (4) We apply the transfer network on \domainB{} to solve \tasktwo{} without the need for annotations.}
	\label{fig:framework}
	\end{figure*}
	
	We wish to start with a practical example of the problem we are trying to solve and how we address it. Let us consider a synthetic and a real domain where we aim to solve the semantic segmentation task. Annotations come for free in the synthetic domain while are rather expensive in the real one. Domain adaptation comes handy for this; however, we wish to go one step further.
	May we pick a closely related task (\eg, depth estimation) where annotations are available in both domains and use it to boost the performance of semantic segmentation on real data?
	To achieve this goal we train deep networks for depth and semantic segmentation on the synthetic domain and learn a mapping function to transform deep features suitable for depth estimation into deep features suitable for semantic segmentation.   
	Then we apply the same mapping function on samples from the real domain to obtain a semantic segmentation model without the need of semantic labels in the real domain. In the remainder of this section, we formalize the \algoname{} framework. 
	
	\subsection{Common Notation}
	We denote with $\mathcal{T}_j$ a generic visual task defined as in \cite{zamir2018taskonomy}. 
	Let us assume $\mathcal{X}^k$ to be the set of samples (\ie, images) belonging to domain \textit{k} and $\mathcal{Y}_j^k$ to be the paired set of annotations for task $\mathcal{T}_j$.
	In our problem we assumes to have two domains, \domainA{} and \domainB{}, and two tasks, \taskone{} and \tasktwo{}. 
	For the two tasks we have complete supervision in \domainA{}, \ie, ${\mathcal{Y}^A_1}$ and ${\mathcal{Y}^A_2}$, but labels only for \taskone{} in  \domainB{}, \ie $\mathcal{Y}^B_1$.
	We assume each task $\mathcal{T}_j$ to be solvable by a deep neural network $N_j$, consisting in a feature encoder $E_j$ and a feature decoder $D_j$, such that $\hat{y}_j=N_j(x)=D_j(E_j(x))$.
	The network is trained on domain $k$ by minimizing a task-specific loss on annotated samples $(x^k,y^k_j)\sim(\mathcal{X}^k,\mathcal{Y}_j^k)$.
	The result of this training is a network trained to solve $\mathcal{T}_j$ using samples from $\mathcal{X}^k$ that we denote as $N_j^k$.

	\subsection{Overview}\label{ss:overview}
	Our work builds on the intuition that if two tasks are related there should be a function $G_{1\rightarrow2}: \mathcal{T}_1 \rightarrow \mathcal{T}_2$ that transfer knowledge among them.
	But what does transferring knowledge actually means? 
	We will show that this abstract concept can be implemented by transferring representations in deep feature spaces. 
	We propose to first train two task specific networks, $N_1$ and $N_2$, then approximate function \trnet{} by a deep neural network that transforms features extracted by $N_1$ into corresponding features extracted by $N_2$ (\ie, $G_{1\rightarrow2}:E_1(x) \rightarrow E_2(x)$).
	We train \trnet{} by minimizing a reconstruction loss on \domainA{}, where we have complete supervision for both tasks, and use it on \domainB{} to solve \tasktwo{} having supervision only for \taskone{}.  
	
	Our method can be summarized into the four steps pictured in \autoref{fig:framework} and detailed in the following sections:
	\begin{enumerate}
		\itemsep0em
		\item
		Learn to solve task \taskone{} on domains \domainA{} and \domainB{}.
		\item
		Learn to solve task \tasktwo{} on domain \domainA{}.
		\item
		Train \trnet{} on domain \domainA{}.
		\item 
		Apply \trnet{} to solve \tasktwo{} on domain \domainB{}.
	\end{enumerate}
	
	\subsection{Solve \taskone{} on \domainA{} and \domainB{}}\label{ss:step1}
	A network $N_1$ can be trained to solve task \taskone{} on domain $\mathcal{X}^k$ by minimizing a task specific supervised loss 
	\begin{equation}
	L_{\mathcal{T}_1}(\hat{y}_1^k,y_1^k);\; \hat{y}_1^k=N_1(x^k).
	\end{equation}
	However, training one network for each domain would likely result in disjoint feature spaces; we, instead, wish to have similar representation to ease  generalization of \trnet{} across domains.
	Therefore, we train a single network, \sourcenet{}, on samples from both domains, \ie, $\mathcal{X}^k=\mathcal{X}^A\cup{\mathcal{X}^B}$.
	Having a common representation ease the learning of a task transfer mapping valid on both domains though training it only on \domainA{}.
	More details on the impact of having common or disjoint networks are reported in \autoref{ss:shared_no_shared}.
	
	\subsection{Solve \tasktwo{} on \domainA{}}
	Now we wish to train a network to solve \tasktwo{}, however, for this task we can only rely on annotated samples from \domainA{}.
	The best we can do is to train a \targetnet{} minimizing a supervised loss
	\begin{equation}
	L_{\mathcal{T}_2}(\hat{y}_2^A,y_2^A);\; \hat{y}_2^A=N_2(x^A).
	\end{equation}
	
	\subsection{Train \trnet{} on \domainA{}}
	We are now ready to train a task transfer network \trnet{} that should learn to remap deep features suitable for \taskone{} into good representations suitable for \tasktwo{}.
	Given \sourcenet{} and \targetnet{} we generate a training set with pairs of features $(E_1^{A\cup{B}}(x^A),E_2(x^A))$ obtained feeding the same input $x^A$ to \sourcenet{} and \targetnet{}.
	We use only samples from \domainA{} for the training set as it is the only domain where we are reasonably sure that the two networks perform well.
	We optimize the parameters of \trnet{} by minimizing the reconstruction error between transformed and target features 
	\begin{equation}
	L_{Tr}= ||G_{1\rightarrow2}(E_1^{A\cup{B}}(x^A)) - E_2^{A}(x^A)))||_2,
	\end{equation}
	At the end of the training \trnet{} should have learned how to remap deep features from one space into the other.
	
	Among all the possible splits $(E, D)$ obtained cutting $N$ at different layers, we select as input for \trnet{} the deepest features, \ie, those at the lowest spatial resolution.
	We make this choice because deeper features tend to be  less connected to a specific domain and more correlated to higher level concepts. 
	Therefore, by learning a mapping at this level we hope to suffer less from domain shift when applying \trnet{} on samples from \domainB{}.
	A more in depth discussion on the choice of $E$ is reported in \autoref{ablation_level}. Additional considerations on the key role of \trnet{} in our framework can be found in the supplementary material.
	
	\subsection{Apply \trnet{} to solve \tasktwo{} on \domainB{}}
	Now we aim to solve \tasktwo{} on \domainB{}.
	We can use the supervision provided for \taskone{} on \domainB{} to extract good image features (\ie, $E^{A\cup{B}}_1(x_B)$ ).
	Then use \trnet{} to transform these features into good features for \tasktwo, and finally decode them through a suitable decoder $D^A_2$.
	The whole system at inference time corresponds to:
	\begin{equation}
	\hat{y}^B_2 = D^A_2(G^A_{1\rightarrow2}(E^{A\cup{B}}_1(x_B)))
	\end{equation}
	
	Thus, thanks to our novel formulation, we can learn through supervision the dependencies between two tasks in a source domain and leverage on them to perform one of the two tasks in a different target domain where annotations are not available.
	%Thanks to our novel formulation,  supervision for a certain task can be used to solve another (related) task on a novel domain without the need of annotations for the target task and domain.
	
	\section{Experimental Settings}\label{settings}
	We describe here the experimental choices made when testing \algoname{}, with additional details provided in the supplementary material due space constraints. 
	
	\textbf{Tasks.}
	To validate the effectiveness of \algoname{}, we select as \taskone{} and \tasktwo{} \emph{semantic segmentation} and \emph{monocular depth estimation}. In the supplementary material, we report some promising results also for other tasks. We minimize a cross entropy loss to train a network for semantic segmentation while we use a $L_1$ regression loss to train a network for monocular depth estimation.
	We choose these two tasks since they are closely related, as highlighted in recent works \cite{ramirez2018geometry, chennupati2019auxnet, Cipolla_2018}, and of clear interest in many practical settings such as, \eg{}, autonomous driving.
	Moreover, as both tasks require a structured output, they can be addressed by a similar network architecture with the only difference being the number of filters in the final layer: as many as the number of classes for  semantic segmentation and just one for depth estimation.
	
	\textbf{Datasets.}
	We consider four different datasets, two synthetic ones, and two real ones.
	We pick synthetic datasets as \domainA{} to learn the mapping across tasks thanks to availability of free annotations.
	We use real dataset as \domainB{} to benchmark the performance of \algoname{} in challenging realistic conditions.
	As synthetic datasets we have used the six video sequences of the Synthia-SF dataset \cite{HernandezBMVC17} (shortened as \synthia{}) and rendered several other sequences with the \carla{} simulator \cite{Dosovitskiy17}. 
	For both datasets, we have split the data into a train, validation, and test set by subdividing them at the sequence level (\ie, we have used different sequences for train, validation, and test). 
	As for the real datasets, we have used images from the \kitti{} \cite{KITTI_2012,KITTI_2015,KITTI_RAW} and  \cityscapes{} \cite{Cordts_2016_CVPR} benchmarks.
	Concerning \kitti{}, we have used the 200 images from the \kitti{} 2012 training set \cite{KITTI_2012} to benchmark depth estimation and 200 images from the \kitti{} 2015 training set with semantic annotations recently released in \cite{Alhaija2018IJCV}.
	As for \cityscapes{}, we have used the validation split to benchmark semantic segmentation and all the images in the training split.
	When training depth estimation networks on \cityscapes{}, following a procedure similar to \cite{tonioni2017unsupervised}   we generate proxy labels by filtering SGM \cite{hirschmuller2005accurate} disparities through confidence measures (left-right check).
	
	\begin{table*}[t]
		\center
		\setlength{\tabcolsep}{2.5pt}
		\scalebox{0.9}{
		\begin{tabular}{lccc|ccccccccccc|cc}
			\toprule
			%&\multicolumn{18}{c}{Per-class mIoU}& & \\
			%&\cline{1-19}&&\\
			&\domainA{} & \domainB{} & Method & \rotatebox{90}{Road} & \rotatebox{90}{Sidewalk} & \rotatebox{90}{Walls} & \rotatebox{90}{Fence} & \rotatebox{90}{Person} & \rotatebox{90}{Poles} & \rotatebox{90}{Vegetation} & \rotatebox{90}{Vehicles} & \rotatebox{90}{Tr. Signs} & \rotatebox{90}{Building}  & \rotatebox{90}{Sky} & \textbf{mIoU} & \textbf{Acc} \\
			\midrule 
			\multirow{2}{*}{\textbf{(a)}}&\synthia{} & \carla{} & Baseline & 63.94 & 54.87 & 15.21 & \textbf{0.03} & 13.55 & 12.78 & \textbf{52.73} & 27.34 & 4.88 & 50.24 & 79.73 & 34.12 & 73.36 \\
			&\synthia{} & \carla{} & \algoname{} & \textbf{73.57} & \textbf{62.58} & \textbf{26.85} & 0.00 & \textbf{17.79} & \textbf{37.30} & 35.27 & \textbf{52.94} & \textbf{17.76} & \textbf{62.99} & \textbf{87.50} & \textbf{43.14} &\textbf{ 80.00} \\ 
			\midrule
			\multirow{2}{*}{\textbf{(b)}}&\synthia{} & \cityscapes{} & Baseline & 6.91 & 0.68 & 0.00 & 0.00 & 2.47 & 9.14 & \textbf{3.19} & \textbf{8.90} & \textbf{0.81} & 25.93 & 26.86 & 7.72 & 28.49 \\
			&\synthia{} & \cityscapes{} & \algoname{} & \textbf{85.77} & \textbf{29.40} & \textbf{1.23} & 0.00 & \textbf{3.72} & \textbf{14.55} & 1.87 & 8.85 & 0.38 & \textbf{42.79} & \textbf{67.06} & \textbf{23.24} & \textbf{64.03}\\
			\midrule
			\multirow{3}{*}{\textbf{(c)}}&\carla{} & \cityscapes{} & Baseline & 71.87 & \textbf{36.53} & 3.99 & \textbf{6.66} & 24.33 & 22.20 & 66.06 & \textbf{48.12} & 7.60 & 60.22 & 69.05 & 37.88 & 74.61 \\
			&\carla{} & \cityscapes{} & \algoname{} & \textbf{76.44} & 32.24 & \textbf{4.75} & 5.58 & \textbf{24.49} & \textbf{24.95} & \textbf{68.98} & 40.49 & \textbf{10.78} & \textbf{69.38} & \textbf{78.19} & \textbf{39.66} & \textbf{76.37} \\
			& - & \cityscapes{} & Oracle & 95.65 & 77.72 & 33.02 & 37.63 & 65.45 & 42.087 & 89.36 & 89.99 & 41.36 & 86.81 & 89.22 & 68.02 & 93.56 \\
			\bottomrule
		\end{tabular}
		}
		%\captionsetup{size=small,skip=0.333\baselineskip}
		\small
		\caption{Experimental results of \depsem{} scenario. Best results highlighted in bold.}
		\label{tab:depth2sem}
	\end{table*}
	
	\textbf{Network Architecture.}
	Each task network is implemented as a dilated ResNet50 \cite{Yu2017} that compresses an image to $1/16$ of the input resolution to extract features.
	Then we use several bilinear up-sample and convolutional layers to regain resolution and get to the final prediction layer. 
	All the layers of the network feature batch normalization.
	We implement the task transfer network (\trnet{}) as a simple stack of convolutional and deconvolutional layers that reduce the input to $1/4$ of the input resolution before getting back to the original scale.
	
	\textbf{Evaluation Protocol.}
	For each test we select two domains (\ie, two datasets,  referred to as \domainA{} and  \domainB{}) and one direction of task transfer, \eg{},  from \taskone{} to \tasktwo{}.
	We will use \semdep{} when mapping features from semantics to depth and \depsem{} when switching the two tasks.
	For each configuration of datasets and tasks we use \algoname{} to train a cross-task network (\trnet{}) following the protocol described in \autoref{ss:overview}, then measure its performance for \tasktwo{} on \domainB{}.
	We compare our method against a \emph{Baseline} obtained training a network with supervision for \tasktwo{} in \domainA{} (\ie, $N_2^A$) and testing it on \domainB{}.
	Moreover, we report as a reference the performance  attainable by a \emph{Oracle} (\ie{}, a network trained with supervision on \domainB{}). 
	
	\textbf{Metrics.}
	Our semantic segmentation networks predict eleven different classes corresponding to those available in the \carla{} simulator plus one additional class for `Sky'.
	To measure performance, we report two different global metrics: pixel accuracy, shortened \emph{Acc.} (\ie, the percentage of pixels with a correct label) and Mean Intersection Over Union, shortened \emph{mIoU} (computed as detailed in \cite{Cordts_2016_CVPR}).
	To provide more insights on per-class gains we also report the \emph{IoU} (intersection-over-union) score computed independently for each class.
	
	When testing the depth estimation task we use the standard metrics described in \cite{eigen2014depth}: Absolute Relative Error (Abs Rel), Square Relative Error (Sq Rel), Root Mean Square Error (RMSE), logarithmic RMSE and three $\delta$ accuracy scores ($\delta_\alpha$ being the percentage of predictions whose maximum between ratio and inverse ratio with respect to the ground truth is lower than $1.25^\alpha$).

	\section{Experimental Results}\label{experiments}
	
		\begin{table*}[t]
		\center
		\scalebox{0.9}{
		\begin{tabular}{lccc|cccc|ccc}
			\toprule
			%\cline{6-9}
			\multicolumn{4}{c|}{} & \multicolumn{4}{c|}{\cellcolor{blue!25}Lower is better}
			& \multicolumn{3}{c}{\cellcolor{LightCyan}Higher is better}\\
			&\domainA{} & \domainB{} & Method & \cellcolor{blue!25}Abs Rel & \cellcolor{blue!25}Sq Rel & \cellcolor{blue!25}RMSE & \cellcolor{blue!25}RMSE log & \cellcolor{LightCyan}$\delta_1$ & \cellcolor{LightCyan}$\delta_2$ & \cellcolor{LightCyan}$\delta_3$\\
			\midrule
			\multirow{2}{*}{\textbf{(a)}}& \synthia{} & \carla{} & Baseline & 0.632 & 8.922 & 13.464 & 0.664 & 0.323 & 0.578 & 0.733\\
			&\synthia{} & \carla{} & \algoname{} & \textbf{0.316} & \textbf{5.485} & \textbf{11.712} & \textbf{0.458} & \textbf{0.553} & \textbf{0.785} & \textbf{0.880}\\
			\midrule
			\multirow{3}{*}{\textbf{(b)}}& \carla{} & \cityscapes{} & Baseline & 0.667 & 13.500 & 16.875 & 0.593 & 0.276 & 0.566 & 0.770  \\
			& \carla{} & \cityscapes{} & \algoname{}  & \textbf{0.394} & \textbf{5.837} & \textbf{13.915} &\textbf{ 0.435} & \textbf{0.337} & \textbf{0.749} & \textbf{ 0.899}\\
			& \cityscapes{} & \cityscapes{} & Oracle & 0.176 & 3.116 & 9.645 & 0.256 & 0.781 & 0.921 & 0.969\\
			\midrule
			\multirow{3}{*}{\textbf{(c)}}&\carla{} & \kitti{} & Baseline &  0.500 & 10.602 & 10.772 &  0.487 &  0.384 &  0.723 &  0.853 \\
			& \carla{} & \kitti{} & \algoname{} & \textbf{0.439} & \textbf{8.263} & \textbf{9.148} & \textbf{0.421} &  \textbf{0.483} &  \textbf{0.788} &  \textbf{0.891} \\
			& - & \kitti{} & Oracle & 0.265 & 2.256 & 5.696 & 0.319 & 0.672 & 0.859 & 0.939 \\
			\bottomrule
		\end{tabular}
		}
		\caption{Experimental results of \semdep{} scenario. Best results highlighted in bold.}
		\label{tab:sem2depth}
	\end{table*}
	
	We subdivide the experimental results in three main sections: in \autoref{ss:dep_sem} and \autoref{ss:sem_dep} we transfer information between semantic segmentation and depth estimation, while in \autoref{ss:dom_adapt} we show preliminary results on the integration of \algoname{} with domain adaptation techniques.
	
	\subsection{Depth to Semantics}\label{ss:dep_sem}

	Following the protocol detailed in \autoref{settings}, we first test \algoname{} when transferring knowledge from the \emph{monocular depth estimation} task to the \emph{semantic segmentation} task, and report the results in \autoref{tab:depth2sem}.
	In this setup, we have  supervision for both tasks in \domainA{} while only for depth estimation in \domainB{}.
	Therefore, for each configuration, we report the results obtained performing semantic segmentation on \domainB{} without any domain-specific supervision.
	
	We begin our investigation by studying the task transfer in a purely synthetic environment, where we can have perfect annotations for all tasks and domains, \ie, we use \synthia{} and \carla{} as \domainA{} and \domainB{}, respectively. 
	The results obtained by \algoname{} and a transfer learning baseline are reported in \autoref{tab:depth2sem}-(a). 
	Comparing the two rows we can clearly see that our method boost performance by \emph{+9.02\%} and \emph{+6,64\%}, for mIoU and Acc, respectively, thanks to the additional knowledge transferred from the depth estimation task.
	
	The same performance boost holds when considering a far more challenging domain transfer between synthetic and real data, \ie, \autoref{tab:depth2sem}-(b) (\synthia{} $\rightarrow$ \cityscapes{}) and \autoref{tab:depth2sem}-(c) (\carla{} $\rightarrow$ \cityscapes{}). 
	In both scenarios, our \algoname{} improves the two averaged metrics (mIoU and Acc.) and most of the per class scores, with gain as large as \emph{+78,86\%} for the Road class in (b). 
	Overall \algoname{} consistently improves predictions for the more interesting classes in autonomous driving scenarios, \eg, Road, Person\dots. 
	The main difficulties for \algoname{} seems to deal with transferring knowledge for classes where  depth estimation is particularly hard (\eg, Vegetation, where synthetic data have far from optimal annotations, or thin structures like Poles and Fences).
	Indeed our model in \autoref{tab:depth2sem}-(c) is still far from the performance obtainable by the same \emph{Oracle} network trained with supervision on \domainB{} for \tasktwo{}. 
	However we wish to point out that in this scenario we do not use any annotation at all on the real \cityscapes{} data, since we automatically generate noisy proxy labels for depth from synchronized stereo frames following \cite{tonioni2017unsupervised}. 
	%Therefore all the results on real data are obtained leveraging only on image samples from the \cityscapes{} domain. 
	
	The top row of \autoref{fig:qualitative} show qualitative results on \cityscapes{} where \algoname{} produces clearly better semantic maps than the baseline network.
	
	\subsection{Semantics to Depth}\label{ss:sem_dep}
	
	Following the protocol detailed in \autoref{settings}, we test \algoname{} when transferring features from \emph{semantic segmentation} to \emph{monocular depth estimation}. 
	In this setup, we have complete supervision for both tasks in \domainA{} and only for semantic segmentation in \domainB{}.
	For each configuration we report in \autoref{tab:sem2depth} the results obtained performing monocular depth estimation on \domainB{} without any domain-specific supervision.
	
\begin{figure*}
		\setlength{\tabcolsep}{1pt}
		\centering
		\begin{tabular}{cccc}
			\tabcolsep0em
			Input & Baseline & \algoname{} & GT \\
			\includegraphics[width=0.23\textwidth]{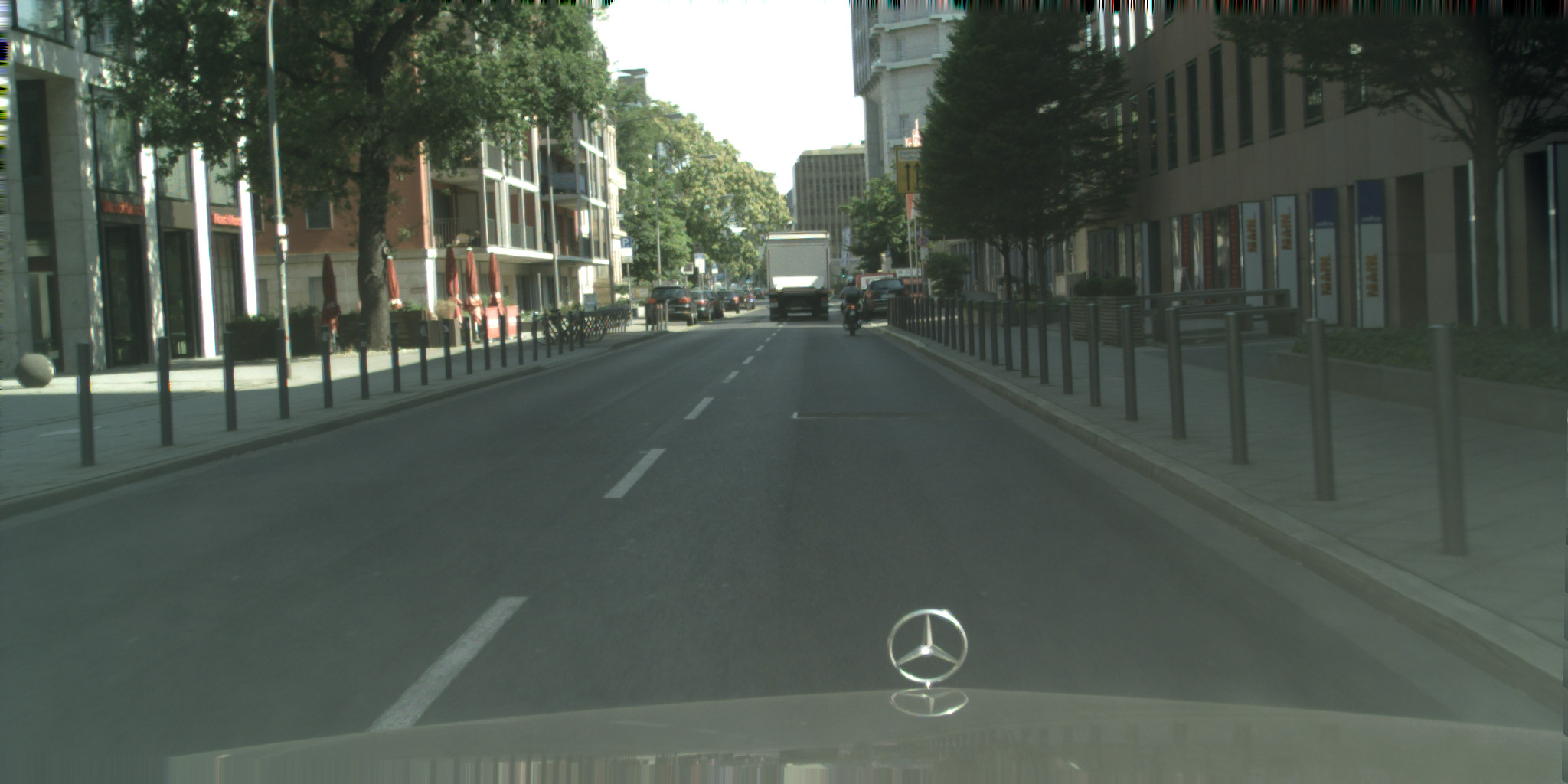} &
			\includegraphics[width=0.23\textwidth]{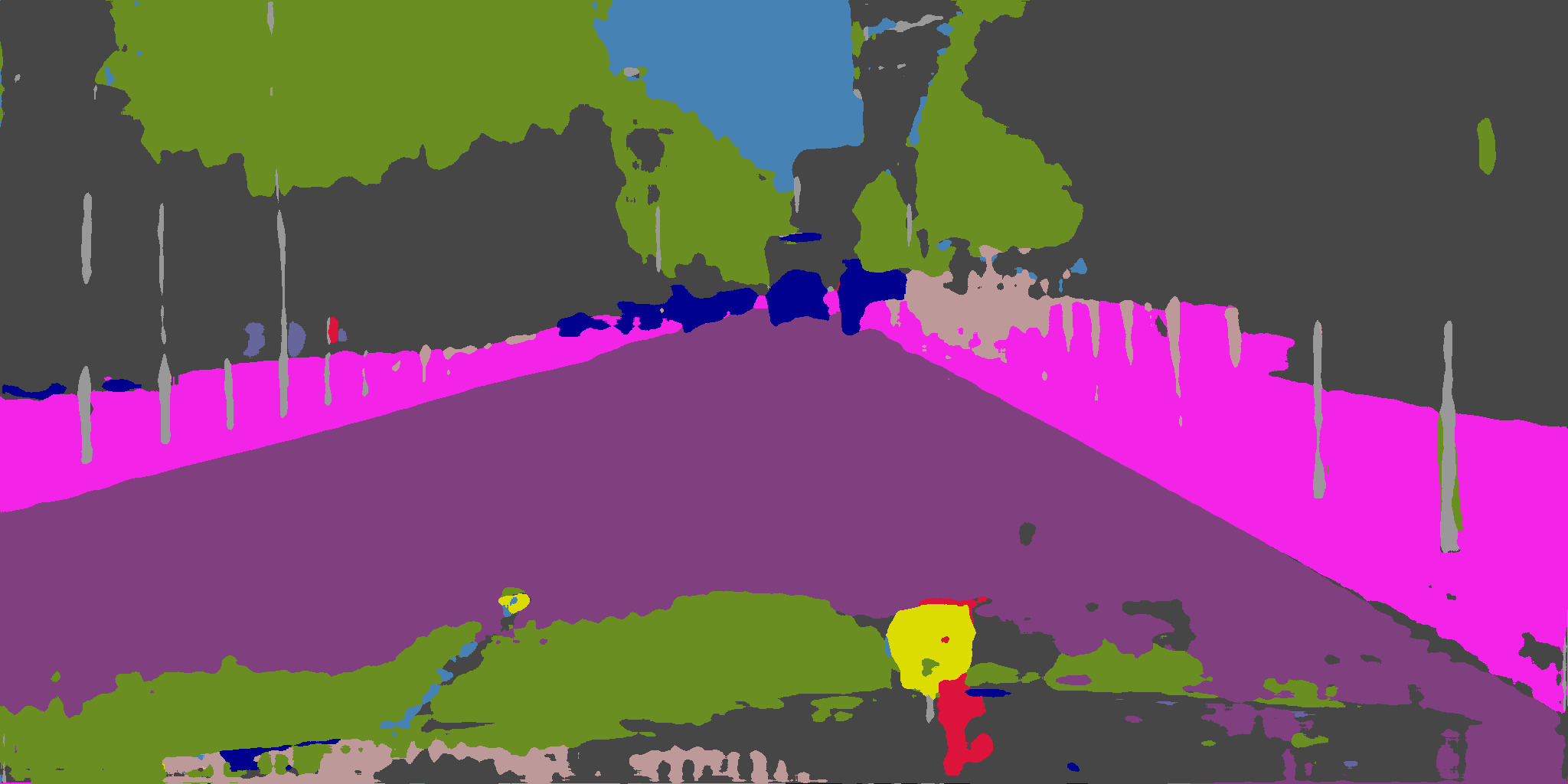} &
			\includegraphics[width=0.23\textwidth]{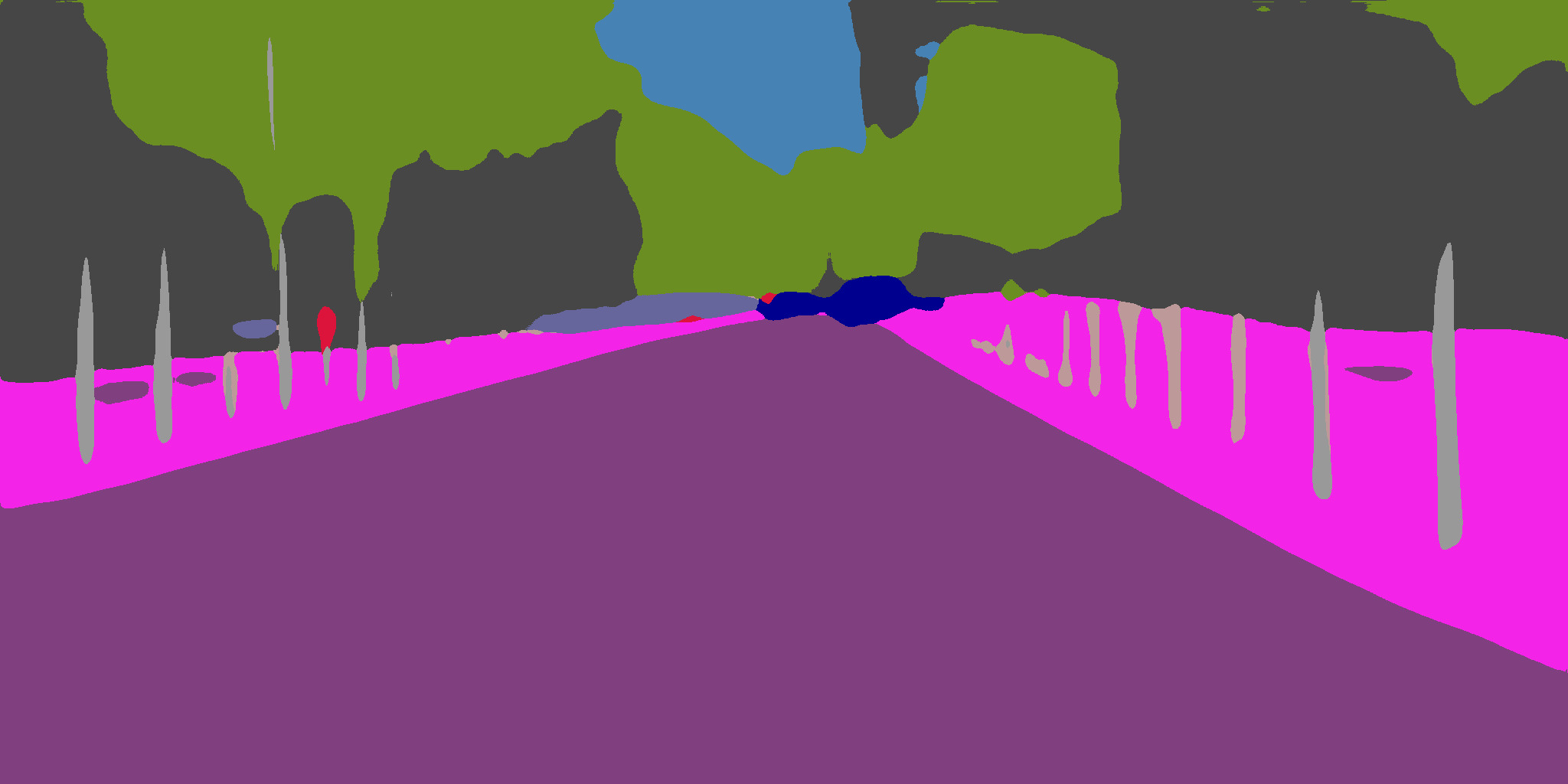} &
			\includegraphics[width=0.23\textwidth]{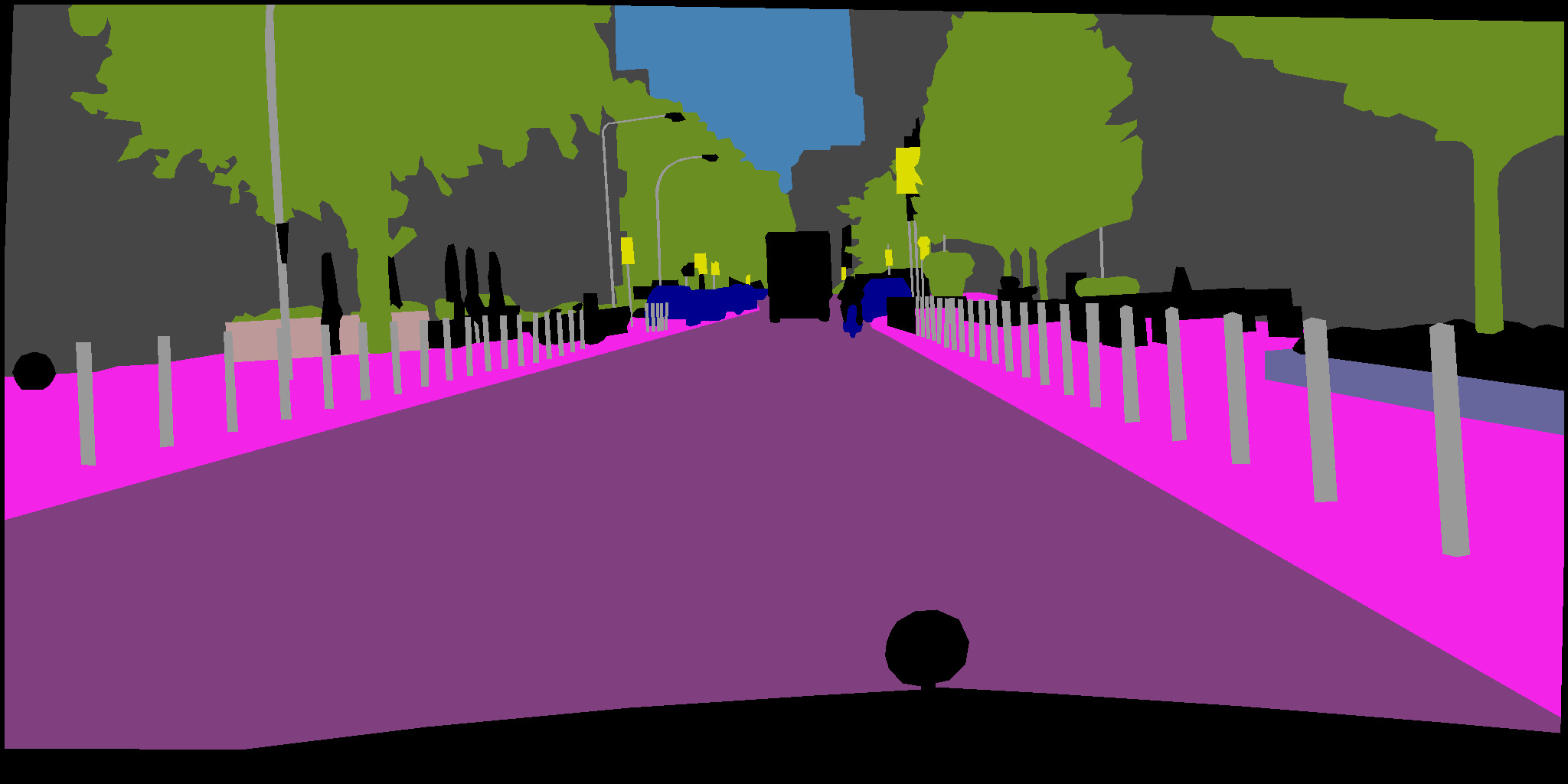} \\
			\includegraphics[width=0.23\textwidth]{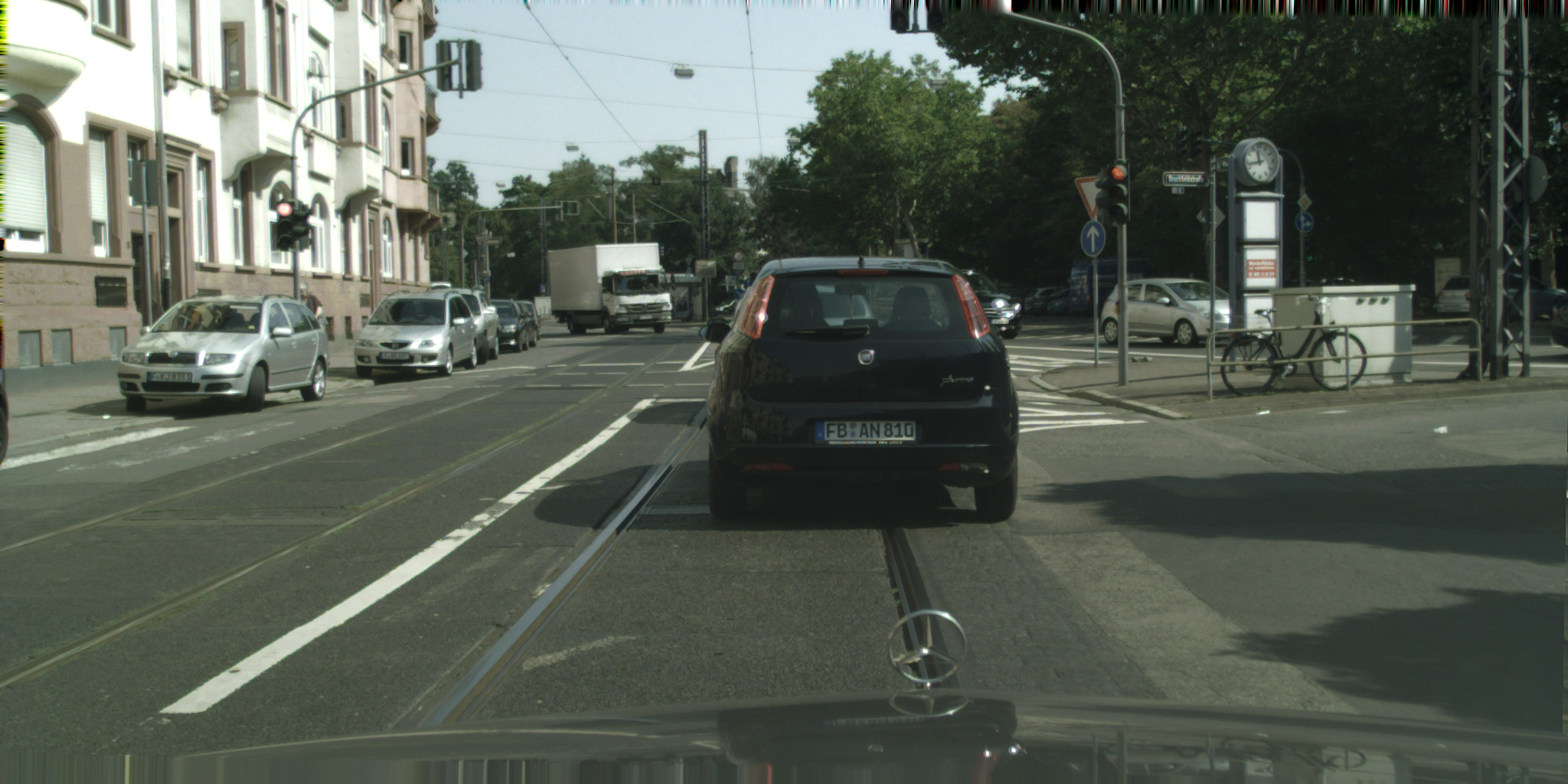} &
			\includegraphics[width=0.23\textwidth]{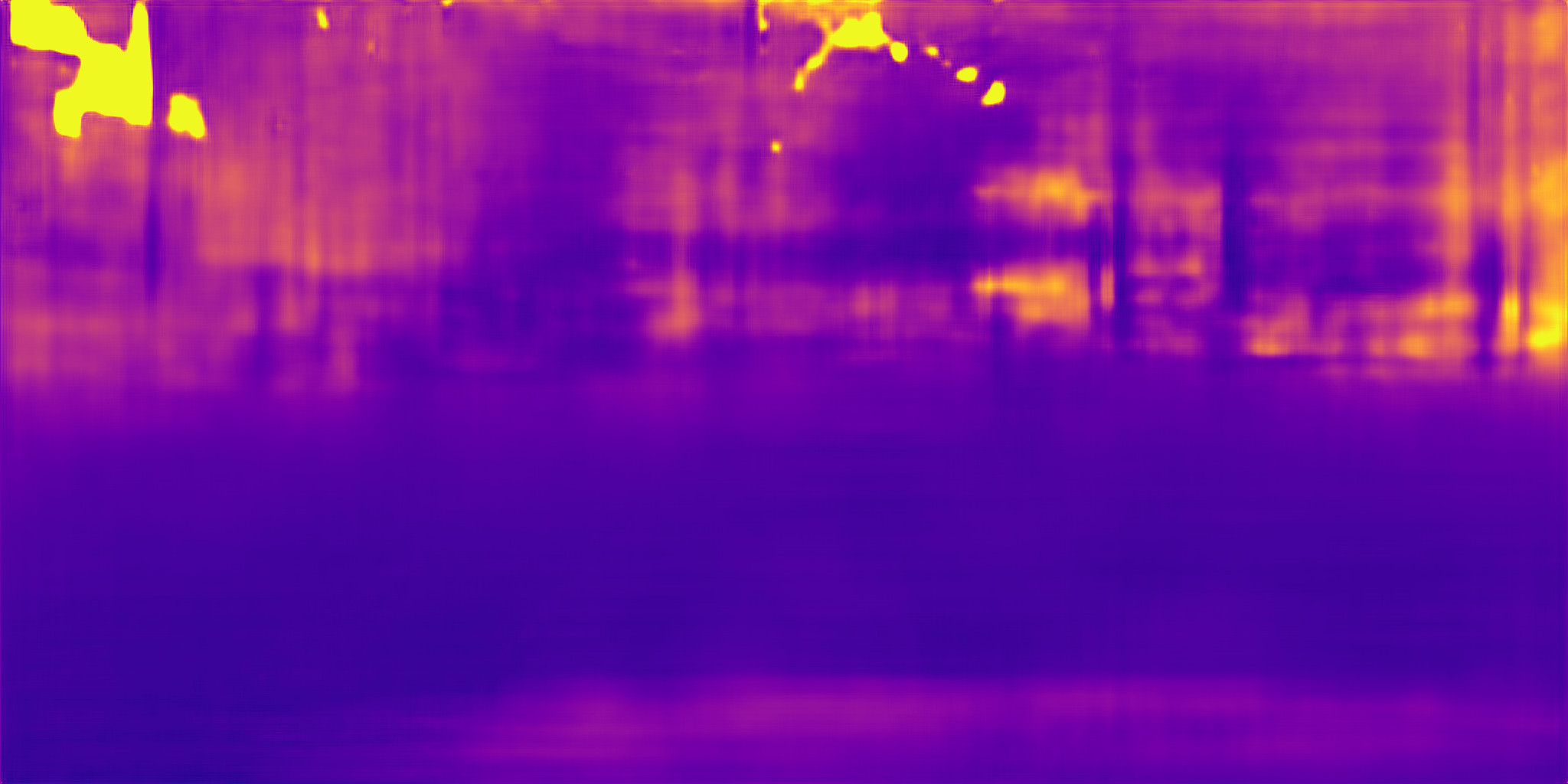} &
			\includegraphics[width=0.23\textwidth]{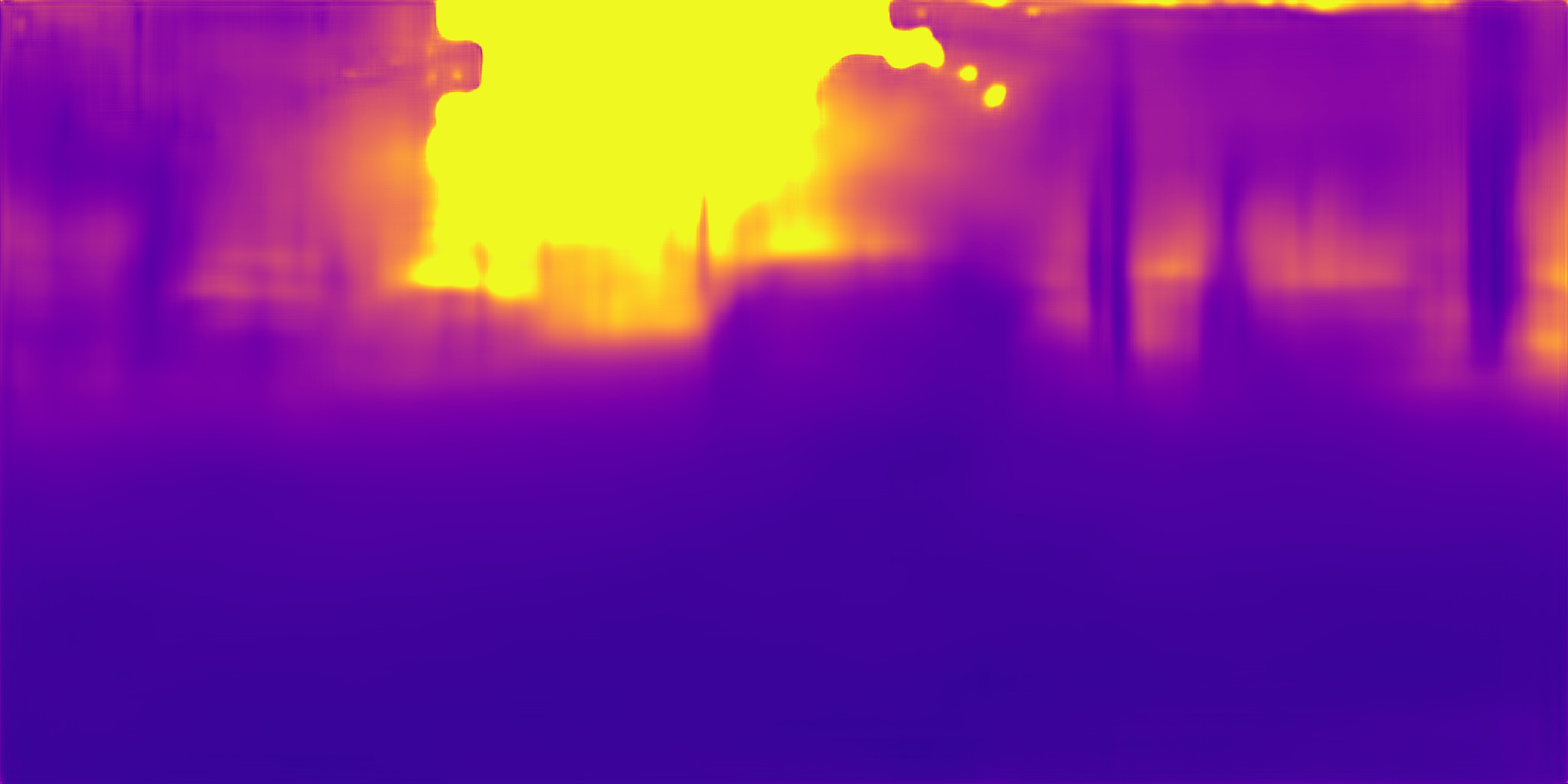} &
			\includegraphics[width=0.23\textwidth]{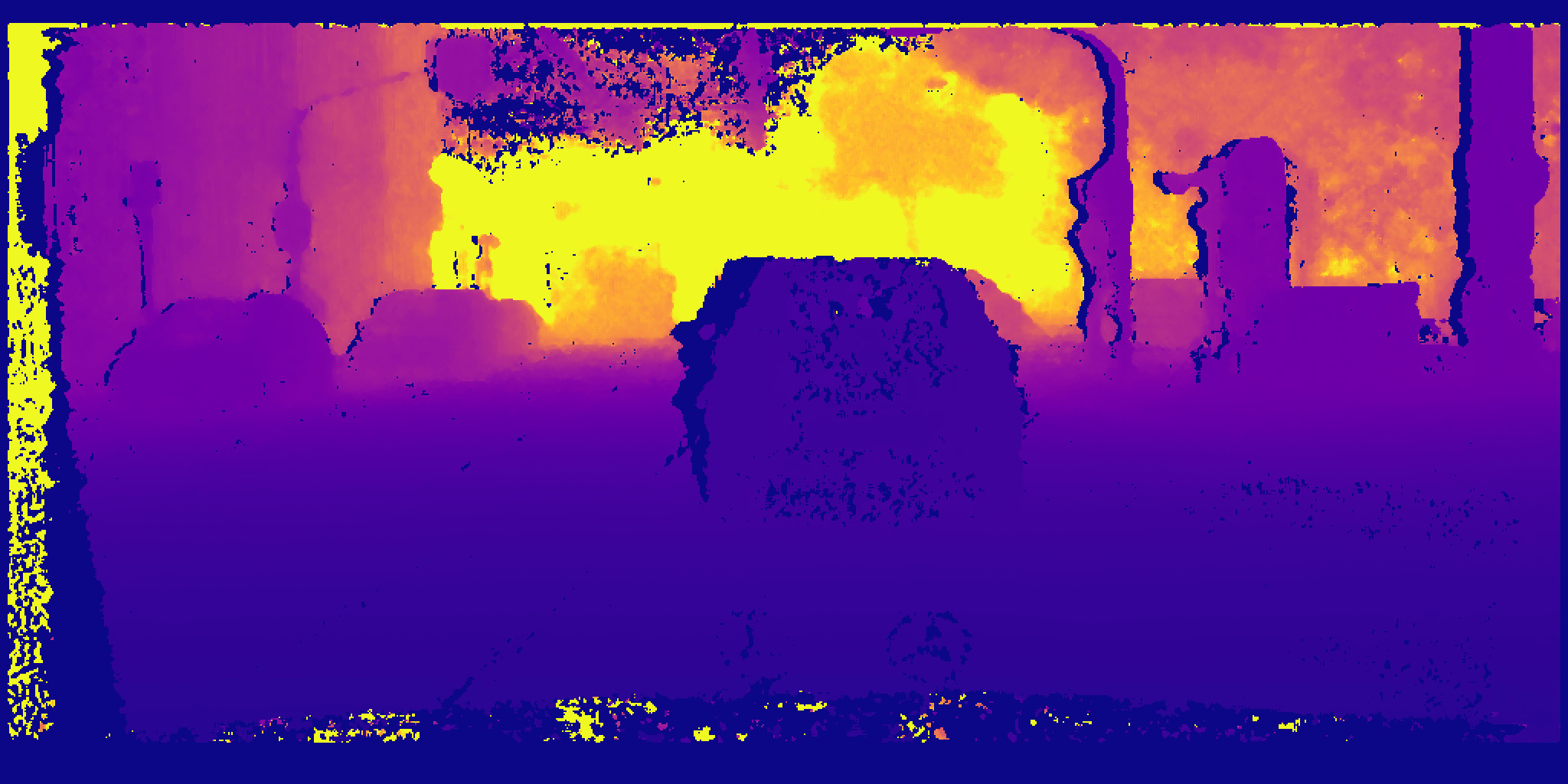} \\
		\end{tabular}
		\caption{Qualitative results for \domainA{}: \carla{} to \domainB{}: \cityscapes{}. First row shows \depsem{} scenario while second row shows \semdep{} setting. From left to right RGB input, baseline predictions, \algoname{} predictions, ground-truth images.}
		\label{fig:qualitative}
	\end{figure*}

	The first pair of rows (\ie, \autoref{tab:sem2depth}-(a)) reports results when transferring knowledge across two synthetic domains.
	The use of knowledge coming from semantic features helps \algoname{} to predict better depths resulting in consistent improvements in all the seven metrics with respect to the baseline.
	The same gains hold for tests concerning real datasets (\ie, \autoref{tab:sem2depth}-(b) with \cityscapes{} and \autoref{tab:sem2depth}-(c) with \kitti{}), where the deployment of \algoname{} always results in a clear advantage against the baseline.
	We wish to point out how on \autoref{tab:sem2depth}-(c) we report a result where \algoname{} use very few annotated samples from \domainB{} (\ie, only the 200 images annotated with semantic labels released by \cite{Alhaija2018IJCV}).
	Comparing \autoref{tab:sem2depth}-(c) to \autoref{tab:sem2depth}-(b) we can see how the low data regime of \kitti{} results in slightly smaller gains, as also testified by the difference among oracle performances in the two datasets. 
	Nevertheless \algoname{} consistently yields improvements with respect to the baseline for all the seven metrics.
	We believe that these results provide some assurances on the effectiveness of \algoname{} with respect to the amount of available data per task.
	Finally, the bottom row of \autoref{fig:qualitative} shows qualitative results on monocular depth estimation on \cityscapes{}: we can clearly observe how \algoname{} provides significant improvements over the baseline, especially on far objects.
	
	\subsection{Integration with Domain Adaptation}\label{ss:dom_adapt}
	
	\begin{table*}[t]
		\center
		\setlength{\tabcolsep}{2.5pt}
		\scalebox{1}{
		\begin{tabular}{lc|ccccccccccc|cc}
			\toprule
			& Method & \rotatebox{90}{Road} & \rotatebox{90}{Sidewalk} & \rotatebox{90}{Walls} & \rotatebox{90}{Fence} & \rotatebox{90}{Person} & \rotatebox{90}{Poles} & \rotatebox{90}{Vegetation} & \rotatebox{90}{Vehicles} & \rotatebox{90}{Tr. Signs} & \rotatebox{90}{Building}  & \rotatebox{90}{Sky} & \textbf{mIoU} & \textbf{Acc} \\
			\hline 
			\textbf{(a)}& Baseline & 71.87 & 36.53 & 3.99 & \textbf{6.66} & 24.33 & 22.20 & 66.06 & 48.12 & 7.60 & 60.22 & 69.05 & 37.88 & 74.61 \\
			\textbf{(b)}& \algoname{} & 76.44 & 32.24 & 4.75 & 5.58 & 24.49 & \textbf{24.95} & 68.98 & 40.49 & \textbf{10.78} & 69.38 & \textbf{78.19} & 39.66 & 76.37 \\
			\textbf{(c)}& CycleGAN & 81.58 & 39.15 & \textbf{6.08} & 5.31 & \textbf{30.22} & 21.73 & \textbf{77.71} & 50.00 & 8.33 & 68.35 & 77.22 & 42.33 & 80.93\\
			\textbf{(d)}& \algoname{} + CycleGAN & \textbf{85.19} & \textbf{41.37} & 5.44 & 3.02 & 29.90 & 24.07 & 71.93 & \textbf{58.09} & 7.53 & \textbf{70.90} & 77.78 & \textbf{43.20} & \textbf{81.92} \\
			\hline
		\end{tabular}}
		\caption{Experimental results of integration with domain adaptation techniques. We show results of \domainA{}: \carla{} to \domainB{}: \cityscapes{} and \depsem{} scenario. Best results highlighted in bold.}
		\label{tab:domain_adaptation_sem}
	\end{table*}
	
	\begin{table*}[t]
		\center
		\scalebox{0.9}{
		\begin{tabular}{lc|cccc|ccc}
			\toprule
			%\cline{6-9}
			\multicolumn{2}{c|}{} & \multicolumn{4}{c|}{\cellcolor{blue!25}Lower is better}
			& \multicolumn{3}{c}{\cellcolor{LightCyan}Higher is better}\\
			& Method & \cellcolor{blue!25}Abs Rel & \cellcolor{blue!25}Sq Rel & \cellcolor{blue!25}RMSE & \cellcolor{blue!25}RMSE log & \cellcolor{LightCyan}$\delta_1$ & \cellcolor{LightCyan}$\delta_2$ & \cellcolor{LightCyan}$\delta_3$\\
			\midrule
			\textbf{(a)}& Baseline &  0.667 & 13.499 & 16.875 & 0.593 & 0.276 & 0.566 & 0.770\\
			\textbf{(b)}& \algoname{} & \textbf{0.394} & \textbf{5.837} & \textbf{13.915} & \textbf{0.435} & \textbf{0.337} & \textbf{0.749} & \textbf{0.899}\\
			\textbf{(c)} & CycleGAN & 0.943 & 27.026 & 21.666 & 0.695 & 0.218 & 0.478 & 0.690  \\
			\textbf{(d)} & \algoname{}+CycleGAN & 0.563 & 10.789 & 15.636 & 0.489 & 0.247 & 0.668 & 0.861 \\
			\bottomrule
		\end{tabular}}
		\caption{Experimental results of comparison and integration with domain adaptation techniques. We show results of \domainA{}: \carla{} to \domainB{}: \cityscapes{} and \semdep{} scenario. Best results highlighted in bold. }
		\label{tab:domain_adaptation_dep}
	\end{table*}
	
	All the results of \autoref{ss:dep_sem} and \autoref{ss:sem_dep} are obtained learning a mapping function across tasks in a domain and deploying it in another one. 
	Therefore, both the transfer network \trnet{} and the baseline we consider, can indeed suffer from domain shift issues.
	Fortunately, the domain adaptation literature provides several different strategies to overcome domain shifts that are complementary to our \algoname{}. 
	We provide here some preliminary results on how the two approaches may be combined together.
	We consider a pixel-level domain adaptation technique, \ie, CycleGan \cite{Zhu_2017_ICCV}, that transforms images from \domainB{} to render them more similar to those from \domainA{}. 
	In \autoref{tab:domain_adaptation_sem}, we report results obtained for a \depsem{} scenario using \carla{} as \domainA{} and \cityscapes{} as \domainB{}. 
	The pixel level domain alignment of CycleGAN (row (c)) proves particularly effective in this scenario, yielding a huge boost when compared to the baseline (row (a)), even greater then the gain granted by \algoname{} (row (b)).
	However, we can see how the best average results (\ie, mIoU and Acc.) can be obtained combining our cross task framework (\algoname{}) with the pixel level domain adaptation provided by CycleGAN (row (d)).
	Considering the scores on single classes, instead, there is no clear winner among the four considered methods, with different algorithms providing higher accuracy for different classes.  
	In \autoref{tab:domain_adaptation_dep} we report results obtained on a \semdep{} scenario using the same pair of domains and the same four methods. 
	Surprisingly, when targeting depth estimation CycleGAN (row (c)) is not as effective as before and actually worsen significantly the performance of the baseline (row (a)).
	Our \algoname{} is instead more robust to the task being addressed and in this scenario can improve the baseline when combined with CycleGAN (row (d)) and obtain the best overall results when applied alone (row (a)). 
	
	\section{Additional Experiments}\label{sec:additional}
	We report additional tests to shine light on some of the design choices made when developing \algoname{}. Moreover, in the supplementary material we propose an experimental study focused on highlighting the importance of \trnet{}, in particular comparing our proposal to an end-to-end multi-task network featuring a shared encoder and two task dependent decoders. 
	
	\subsection{Study on the Transfer Level}\label{ablation_level}

	\begin{figure}
		\centering
		\includegraphics[width=0.45\textwidth]{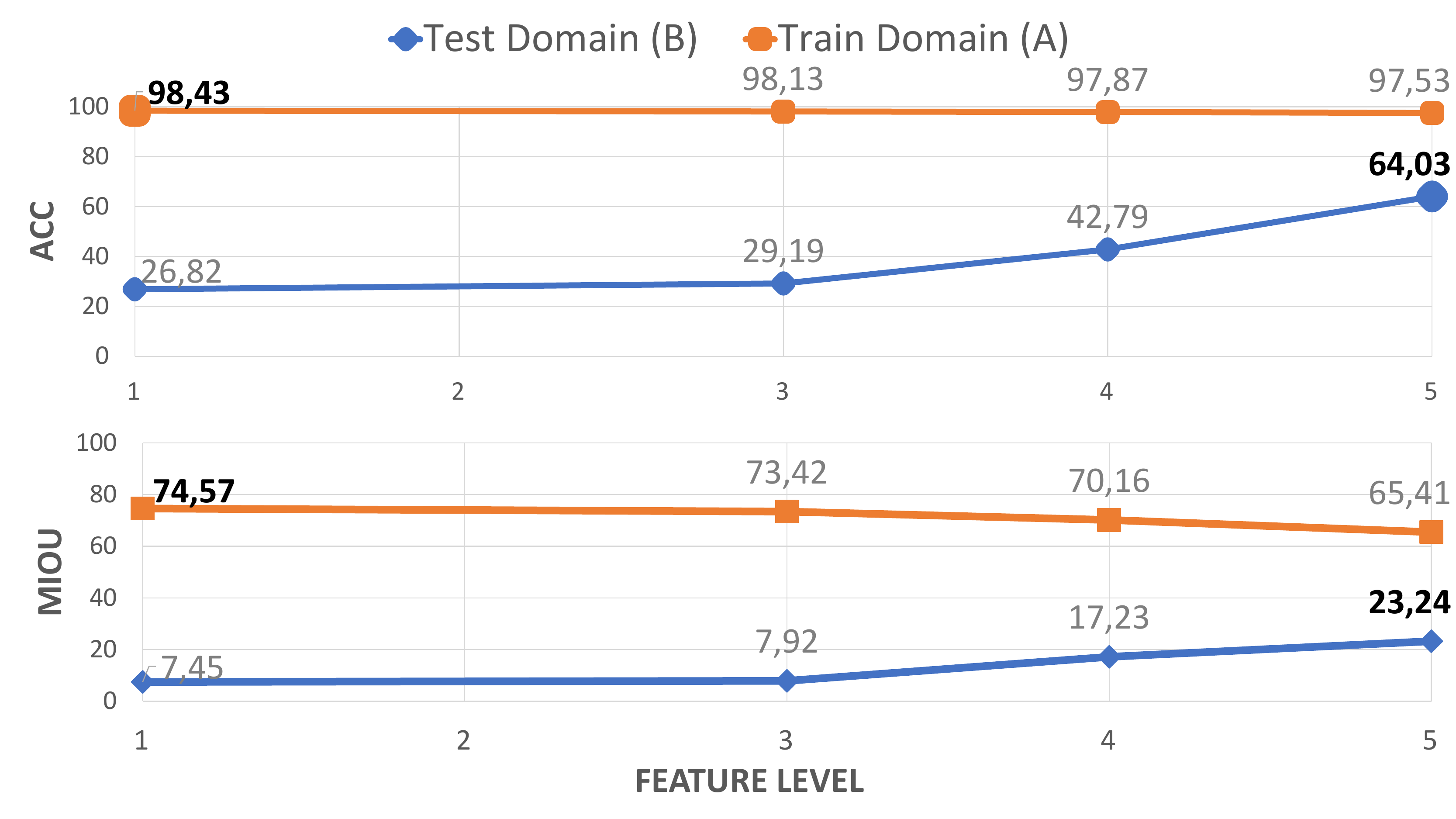} 
		\caption{Study on feature level for task transfer from \synthia{} to \cityscapes{} and \depsem{} scenario. Deeper levels correspond to higher generalization performances.}
		\label{fig:ablationlevels}
	\end{figure}
	
	For all the previous tests we split $N$ between $E$ and $D$ at the layer corresponding to the lowest spatial resolution.
	We pick this split based on the intuition that deeper layers yield more abstract representations, thus less correlated to specific domain information, while lower level features are more domain dependent.
	Therefore, learning \trnet{} between shallower layers should lead to less generalization ability across domains. 
	To validate this intuition, we run experiments aimed at measuring performance for the \depsem{} scenario (\synthia{} $\rightarrow$ \cityscapes{}) when varying the network layer at which we split $N$ into $E$ and $D$.
	We consider four different feature levels corresponding to residual blocks at increasing depth in ResNet50.
	For each  of them we train a transfer network on domain \domainA{} and then measure \textit{mIoU} and \textit{Acc.} testing on unseen images from \domainA{} (\ie{} $D_2^A(G^A_{1\rightarrow2}(E_1^{AuB}(x_A)))$) and \domainB{} (\ie{} $D_2^A(G^A_{1\rightarrow2}(E_1^{AuB}(x_B)))$). The results are plotted  in \autoref{fig:ablationlevels}.
	
	Considering \textit{Acc.} (top plot) we can see how in-domain performance are almost equivalent at the different feature levels (orange line), while cross-domain performance increase when considering deeper feature levels (blue line).
	This pattern is even more pronounced when considering \textit{mIoU} (bottom plot), where in-domain performance actually decreases alongside with deeper feature, whilst cross-domain performance increases. 
	These results validate our intuition that deeper features are less domain specific and may lead to better generalization to unseen domains.
	
	\subsection{Shared vs Non-Shared $N_1$}\label{ss:shared_no_shared}
	
	\begin{table}
		\center
		\scalebox{0.93}{
		\begin{tabular}{cccc}
			\toprule
			\textbf{Shared} & \textbf{Domain} & \textbf{mIoU} & \textbf{Acc.}\\
			\midrule
			\xmark & \domainA{} & 61.73 & 97.02 \\
			\cmark & \domainA{} & 65.41 & 97.53 \\
			\midrule
			\xmark & \domainB{} &  6.42 & 29.36 \\
			\cmark & \domainB{} & 23.24 \textbf{(+16.82)} & 64.03 \textbf{(+34.67)} \\
			\bottomrule
		\end{tabular}
		}
		\caption{Study on Shared vs Non-Shared \sourcenet{}. We show a \domainA{}: \synthia{} to \domainB{}: \carla{} and \depsem{} scenario. Performance improvement highlighted in bold.}
		\label{tab:ablation_encoder}
	\end{table}
	
	Throughout this work we have always trained a single network for \taskone{} with samples from \domainA{} and \domainB{}.
	The rationale behind this choice is to have a single feature extractor for both domains such that \trnet{} trained only on samples from \domainA{} would be able to generalize well to samples from \domainB{} as they are sampled from a similar distribution.
	
	Here we experimentally validate this intuition by comparing a shared \sourcenet{} against the use of two separate networks, one trained on samples from \domainA{} ($N_1^A$) and the other with samples from \domainB{} ($N_1^B$).
	We consider a \depsem{} scenario where we use \synthia{} as domain \domainA{} and \cityscapes{} as \domainB{}. In \autoref{tab:ablation_encoder} we report the \textit{mIoU} and \textit{Acc.} achieved on unseen samples from the two domains.
	On the training domain \domainA{} both methods are able to obtain good results, slightly better for the shared network, probably thanks to the higher variety of data used for training. 
	However, when moving to the completely different domain \domainB{}, it is clear that maintaining the same feature extractor is of crucial importance to be able to use the same \trnet{}. 
	This test suggests the interesting findings that feature extracted by the exact same network architecture trained for the exact same tasks in two different domains are quite different.
	Therefore to correctly apply \trnet{} we need to take into account these difficulties.  
	
	\subsection{Batch Normalization}\label{ablation_bn}
	
	\begin{table}
		\center
		\scalebox{0.93}{
		\begin{tabular}{cccc}
			\toprule
			\textbf{Batchnorm} & \textbf{Domain} & \textbf{mIoU} & \textbf{Acc.}\\
			\midrule
			\xmark & \domainA{} & 72.48 & 98.09 \\
			\cmark & \domainA{} & 65.41 & 97.53 \\
			\midrule
			\xmark & \domainB{} & 22.75 & 58.29 \\
			\cmark & \domainB{} & 23.24 \textbf{(+0.49)} & 64.03 \textbf{(+5.74)} \\
			\bottomrule
		\end{tabular}
		}
		\caption{Ablation Study on Batch Normalization. We show a \domainA{}: \synthia{} to \domainB{}: \cityscapes{} and \depsem{} scenario. Performance improvement highlighted in bold.}
		\label{tab:ablation_bn}
	\end{table}
	
	We investigate the impact on performance of using task networks with or without batch normalization layers \cite{Ioffe2015}.
	Our intuition is that the introduction of batch normalization yields more similar features across domains and smaller numerical values, making the training of \trnet{} easier and numerically more stable.
	In \autoref{tab:ablation_bn} we report results for the \depsem{} scenario when employing \synthia{} as \domainA{} and \cityscapes{} as \domainB{}. 
	As expected, batch normalization yields representations more similar between domains, thus leading to better generalization performances on \domainB{}.
	Counter-intuitively, we also notice that results on \domainA{} are worse with batch normalization, perhaps due to mapping  features from \taskone{} to \tasktwo{} being harder when these 
	lay within a more constrained space. 
	
	\section{Conclusion and Future Works}
	We have shown that it is possible to learn a mapping function to transform deep representations suitable for specific tasks into others amenable to different ones.
	Our methods allows for leveraging on easy to annotate domains to solve tasks in scenarios where annotations would be costly.
	We have shown promising results obtained by applying our framework to two tasks (semantic segmentation and monocular depth estimation) and four different domains (two synthetic domains and two real domains). In future work, we plan  to investigate on the effectiveness and robustness of our framework when addressing other tasks. In this respect, \emph{Taskonomy} \cite{zamir2018taskonomy} may guide us in identifying tightly related visual tasks likely to enable effective transfer of learned representations. 
	We have also shown preliminary results concerning how our framework may be fused with standard domain adaptation strategies in order to further ameliorate performance. We believe that finding the best strategy to fuse the two worlds is a novel and exciting research avenue set forth by our paper.  
	
	{\small
		\bibliographystyle{ieee_fullname}
		\bibliography{egbib}
	}

	

\section*{Supplementary material}

# Supplementary material for Learning Across Tasks and Domains

Pierluigi Zama Ramirez, Alessio Tonioni, Samuele Salti, Luigi Di Stefano
Department of Computer Science and Engineering (DISI)
University of Bologna, Italy
{`pierluigi.zama, alessio.tonioni, samuele.salti, luigi.distefano `}@unibo.it

## 1. Additional Experimental Results

We report here additional experiments to asses the contribution of the different components of AT/DT. In Sec. 1.1 we conduct a study on the performance achievable on the training domain $\mathcal{A}$ with $G_{1\to 2}$. In Sec. 1.2 we show ablation studies to confirm the key role of $G_{1\to 2}$ in our formulation. In Sec. 1.3 we perform tests considering different approaches to build a shared feature representation for $N_1^{A\cup B}$. In Sec. 1.4 we provide additional details and results on the integration of AT/DT with existing domain adaptation techniques. In Sec. 1.5 we report qualitative results using normal estimation as target task ($\mathcal{T}_2$). Finally, in Sec. 2 and Sec. 3 we provide additional details about the training and evaluation processes.

### 1.1. Train domain performance of $G_{1\to 2}$

Our framework has to overcome two nuisances to effectively address the lacking of supervision in the target task and domain: translation of features between tasks and change of domain. In this section, we are interested in isolating the impact of the first nuisance, which will also provide some hints on the importance of the second one. In other words, we are trying to answer the question: *How well are we effectively learning to translate deep representations?*

To focus only on the effectiveness in transferring representations, we consider a test set of images from $\mathcal{A}$ and compare AT/DT and $N_2^A$ (the network trained on domain $\mathcal{A}$ for $\mathcal{T}_2$). As the test data are sampled from the same domain as the training data, we do not have errors due to the domain shift and can use the gap in performance between the two algorithms as a measure of the effectiveness of our framework in transferring representations. As we wish to evaluate both semantic segmentation and depth estimation, we select the Synthia domain as $\mathcal{A}$, for which we have all labels available, and Cityscapes as $\mathcal{B}$. In Tab. 1 we report the results when transferring deep representations in the $Dep. \to Sem.$ scenario, while in Tab. 2 in the $Sem. \to Dep.$ scenario.

Tab. 1 shows how transferring deep representations from $\mathcal{T}_1$ to $\mathcal{T}_2$ with AT/DT results in a small loss in performance compared to $N_2^A$. In particular, the largest performance drops are related to classes dealing with small objects, like 'Fence', 'Poles' and 'Traffic Sign', that might get lost transferring features at the smallest spatial resolution in the network. These results suggest that a multi-scale transfer strategy would be a direction worth exploring in future work to better recover small details upon transferring representations. Nevertheless, the comparisin between the final pixel accuracy (Acc.) highlights that AT/DT loses only 1% though relying on a feature extractor trained for a different task.

In Tab. 2 AT/DT obtains again performance close to $N_2^A$. For some metrics, it even delivers better performance than $N_2^A$. This somewhat surprising result can be explained by the difference between the training sets: AT/DT uses as feature extractor $N_1^{A\cup B}$, which has been trained with samples from both $\mathcal{A}$ and $\mathcal{B}$, *i.e.* with a larger and more varied training set than that used by $N_2^A$. Therefore, the encoder of $N_1^{A\cup B}$ might learn a more general feature extractor than that of $N_2^A$, this resulting in better performance when applied on unseen data. AT/DT can successfully leverage on this better feature extractor and obtain slightly better performance when transferring them to $\mathcal{T}_2$.

The same reasoning may be applied to the results of Tab. 1. However, in this case, the shared encoder of $N_1^{A\cup B}$ has been partially trained with noisy ground truth depth labels on samples from $\mathcal{B}$. The introduction of noise in the training process might harm the learning of $N_1^{A\cup B}$ and explain the small gap in performance. Moreover, as stated above, due to the transferring of features at low resolution, AT/DT might struggle to transfer small image structures (*e.g.*, 'poles', 'traffic sign'...). However, wrong predictions on this kind of small structures do not arm much the depth estimation metrics (*i.e.*, few pixels are considered), though they have a larger impact on the mIoU metric considered for semantic segmentation. Finally, as stated in [2], the advantages yielded by semantic information to depth estimation are larger than the gains attainable going in the other direction, thus motivating the slight difference in performance across the two scenarios.

1

| $\mathcal{A}$ | Method | Road | Sidewalk | Walls | Fence | Person | Poles | Vegetation | Vehicles | Tr. Signs | Building | Sky | mIoU | Acc |
|---|---|---|---|---|---|---|---|---|---|---|---|---|---|---|
| Synthia | $N_2^A$ | **99.23** | **87.16** | **92.67** | **28.62** | **48.53** | **63.54** | **85.02** | **88.92** | **52.67** | **96.91** | **98.39** | **76.52** | **98.45** |
| Synthia | AT/DT | 98.34 | 76.09 | 84.99 | 1.06 | 29.25 | 45.57 | 80.15 | 85.72 | 25.31 | 95.53 | 97.45 | 65.41 | 97.53 |

Table 1: Experimental results of $Dep. \rightarrow Sem.$ scenario using as domain $\mathcal{A}$ the Synthia dataset. Best results highlighted in bold.

| | | Lower is better | | | | Higher is better | | |
|---|---|---|---|---|---|---|---|---|
| $\mathcal{A}$ | Method | Abs Rel | Sq Rel | RMSE | RMSE log | $\delta_1$ | $\delta_2$ | $\delta_3$ |
| Synthia | $N_2^A$ | 0.138 | **1.212** | **4.759** | 0.825 | **0.864** | 0.952 | 0.970 |
| Synthia | AT/DT | **0.135** | 1.271 | 5.061 | **0.634** | 0.863 | **0.958** | **0.977** |

Table 2: Experimental results of $Sem. \rightarrow Dep.$ scenario using as domain $\mathcal{A}$ the Synthia dataset. Best results highlighted in bold.

Overall, the results reported in Tab. 2 and Tab. 1 show that our framework is indeed learning to transfer deep representations effectively and that it is possible to approximate $G_{1\rightarrow 2}$ by a neural network like that we propose in this work. This is further validated in Fig. 1, where we report two t-SNE[3] plots of deep features extracted by $N_1^{A\cup B}$ (in pink), $N_2^A$ (in blue) alongside with the features transformed by $G_{1\rightarrow 2}$ (in red). All features are computed on image samples from the test set described above, *i.e.* samples unseen at training time. Therefore, $G_{1\rightarrow 2}$ takes as input pink points and produces red points that should be as close as possible to the blue points. Indeed, the two plots show how our task transfer network can successfully produce features suitable for $\mathcal{T}_2$.

### 1.2. Importance of $G_{1\rightarrow 2}$

We report results of additional tests to further assess the importance of $G_{1\rightarrow 2}$ in our cross tasks and domains adaptation. Purposely, we consider a single network made out of one encoder, $E_{1,2}^{A\cup B}$ and two decoders, $D_1^{A\cup B}$ and $D_2^A$. $D_1^{A\cup B}$ is trained with samples from $\mathcal{A}$ and $\mathcal{B}$ for $\mathcal{T}_1$. $D_2^A$ is trained with samples from $\mathcal{A}$ for $\mathcal{T}_2$. Finally, $E_{1,2}^{A\cup B}$ is trained together with the two heads with both tasks and domains. Therefore we consider a single feature extractor which yields a shared representation for both tasks and domains without the need to learn a transfer function between tasks. We will refer to this configuration as the *No Transfer* setting.

We evaluate *No Transfer* for both $Dep. \rightarrow Sem.$ and $Sem. \rightarrow Dep.$ settings from Carla to Cityscapes and compare it to AT/DT and the transfer learning baseline of the main paper. Tab. 3 and Tab. 4 report results for $Dep. \rightarrow Sem.$ and $Sem. \rightarrow Dep.$ settings respectively.

For $Sem. \rightarrow Dep.$ our method outperforms *No Transfer* for all metrics, and indeed this alternative is even worse than the baseline for Sq. Rel. and RMSE. On the other hand, for $Dep. \rightarrow Sem.$ our method achieves better performances in the majority of the classes and for the mIoU, while *No Transfer* provides the best pixel accuracy. We ascribe this result to *No Transfer* providing the highest IoU for the *road* class, which represents the vast majority of pixels in an autonomous driving scenario. However this good performance does not translate to other classes such that *No transfer* achieves the worst mIoU, even less than the baseline. These results confirm the importance of learning a mapping function (*e.g.*, $G_{1\rightarrow 2}$) between features to transfer representations between tasks.

### 1.3. Shared Decoder and Separate Encoders for $N_1$

In Sec. 6.2 we highlighted how learning a common representation for $\mathcal{T}_1$ is crucial to learn a transfer function which generalize across domains. In this additional test we show that to learn a good shared representation across domains for one task, we need to share both encoders and decoders in $N_1^{A\cup B}$. For this reason we train a different version of $N_1^{A\cup B}$ with a shared decoder but two encoders, one trained only on $\mathcal{A}$ and the other only on $\mathcal{B}$. Tab. 5 compares this architecture to AT/DT for Synthia to Cityscapes in the $Dep. \rightarrow Sem.$ scenario. Indeed training a shared encoder allows the representation to be more closely related resulting in better performance.

### 1.4. Integration with Domain Adaptation

We report here additional details on how we have used CycleGAN [5] to address domain adaptation.

We train CycleGAN to transform images from Carla ($\mathcal{A}$)

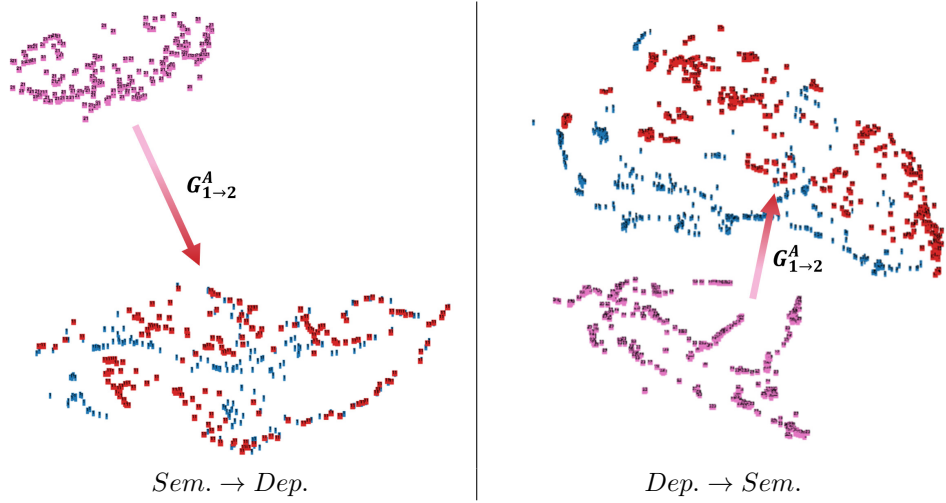

Figure 1: t-SNE [3] plots of deep features computed on $\mathcal{A}$. Pink denotes the features extracted for $\mathcal{T}_1$, i.e. $E_1^{A \cup B}(x_a)$. Blue features extracted for $\mathcal{T}_2$, i.e. $E_2^A(x_a)$. Red the prediction obtained by the feature transfer network $G_{1 \to 2}(E_1^{A \cup B}(x_a))$. Therefore, the red points are the transformations of the pink points according to $G_{1 \to 2}$. With an ideal $G_{1 \to 2}$ red and blue points would perfectly overlap, here we can see that unfortunately this is not the case. Nevertheless our transfer function successfully transform pink features to make them closer to blue ones.

| | $\mathcal{A}$ | $\mathcal{B}$ | Method | Road | Sidewalk | Walls | Fence | Person | Poles | Vegetation | Vehicles | Tr. Signs | Building | Sky | mIoU | Acc |
|---|---|---|---|---|---|---|---|---|---|---|---|---|---|---|---|---|
| (c) | Carla | Cityscapes | Baseline | 71.87 | **36.53** | 3.99 | **6.66** | 24.33 | 22.20 | 66.06 | **48.12** | 7.60 | 60.22 | 69.05 | 37.88 | 74.61 |
| (c) | Carla | Cityscapes | No Transfer | **84.82** | 33.15 | 1.00 | 1.79 | 6.30 | 14.26 | **69.91** | 40.32 | 1.84 | 65.67 | 73.49 | 35.69 | **79.53** |
| (c) | Carla | Cityscapes | AT/DT | 76.44 | 32.24 | **4.75** | 5.58 | **24.49** | **24.95** | 68.98 | 40.49 | **10.78** | **69.38** | **78.19** | **39.66** | 76.37 |

Table 3: Experimental results of $Dep. \to Sem.$ scenario. Best results highlighted in bold.

to Cityscapes ($\mathcal{B}$) and vice-versa. The network is trained using the original author implementation[1] for 200k steps on random image crops of $400 \times 400$ pixels. We use the same hyper-parameters settings as proposed in the original paper.

Once trained, we transform the Cityscapes dataset into the Carla style generating a new $CityscapesLikeCarla$ dataset which we will call $\mathcal{B}like\mathcal{A}$ domain (see Fig. 2). The baseline is then obtained by testing $N_2^A$ with the validation set of $\mathcal{B}like\mathcal{A}$. To integrate AT/DT with CycleGAN, we train a $N_1^{A \cup \{Blike A\}}$ on both $\mathcal{A}$ and $\mathcal{B}like\mathcal{A}$ at step 1 of AT/DT. Then, at step 4, to infer the predictions for $\mathcal{T}_2$ on $\mathcal{B}$, we employ the validation set of $\mathcal{B}like\mathcal{A}$ as done for the baseline. To summarize we train the shared source network on samples obtained from $\mathcal{A}$ and $\mathcal{B}like\mathcal{A}$, then we test all networks on the test set of $\mathcal{B}like\mathcal{A}$ (i.e., Cityscapes images transformed to look like those from Carla).

In Fig. 3 we show some qualitative results obtained when combining AT/DT together with the pixel level domain adaptation obtained through CycleGAN. Comparing the results in the $Sem. \to Dep.$ scenario (first row) with those obtained in a $Dep. \to Sem.$ scenario (second row) we can see how CycleGAN is very effective when targeting the semantic segmentation tasks, much less effective when targeting a depth estimation task. AT/DT, instead, consistently produce better predictions than the baseline in both the considered tasks.

### 1.5. Additional tasks

In Fig. 4 we report additional qualitative results when using as $\mathcal{T}_1$ semantic segmentation and as $\mathcal{T}_2$ normal estimation, with Carla as $\mathcal{A}$ and Cityscapes as $\mathcal{B}$. The results confirm the findings of the semantic to depth scenario, with AT/DT producing clearly better prediction than the baseline network. We report only qualitative results due to the lack of annotations to validate normal estimation on the real Cityscapes data.

---

[1] https://github.com/junyanz/pytorch-CycleGAN-and-pix2pix

| | $\mathcal{A}$ | $\mathcal{B}$ | Method | Lower is better | | | | Higher is better | | |
|---|---|---|---|---|---|---|---|---|---|---|
| | | | | Abs Rel | Sq Rel | RMSE | RMSE log | $\delta_1$ | $\delta_2$ | $\delta_3$ |
| **(b)** | Carla | Cityscapes | Baseline | 0.667 | 13.500 | 16.875 | 0.593 | 0.276 | 0.566 | 0.770 |
| **(b)** | Carla | Cityscapes | No Transfer | 0.615 | 17.578 | 19.924 | 0.533 | 0.284 | 0.646 | 0.845 |
| | Carla | Cityscapes | AT/DT | **0.394** | **5.837** | **13.915** | **0.435** | **0.337** | **0.749** | **0.899** |

Table 4: Experimental results of $Sem. \rightarrow Dep.$ scenario. Best results highlighted in bold.

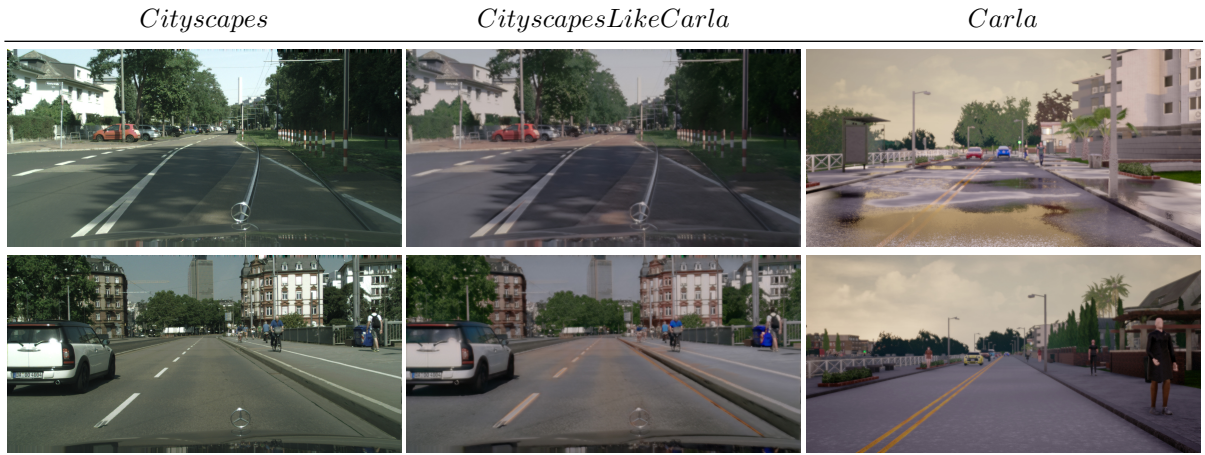

Figure 2: Images obtained applying CycleGAN to make Cityscapes samples similar to those of Carla. From left to right: samples from Cityscapes, corresponding image from $CityscapesLikeCarla$ obtained by CycleGAN, similar samples from Carla

| Shared Encoders | mIoU | Acc. |
|---|---|---|
| ✗ | 11.55 | 56.79 |
| ✓ | 23.24 **(+11.69)** | 64.03 **(+7,24)** |

Table 5: Study on Shared Decoder with Non Shared Encoders for $N_1^{A \cup B}$. We show a $\mathcal{A}$: Synthia to $\mathcal{B}$: Carla and $Dep. \rightarrow Sem.$ scenario. Performance improvement highlighted in bold.

## 2. Details on the training process

Our task networks consist of a ResNet50 as encoder and a stack of 3 series of bilinear upsampler followed by one convolution as the decoder. Our ResNet50 use dilated convolution with rate 2 and 4 in the last two residual blocks, similarly to DRN [4]. We trained our $N_1^{A \cup B}$ and $N_2^A$ until the loss stabilizes with batch size 8 and crop 512x512.

We use Adam [1] as optimizer with a linear decaying learning rate $10^{-4}$ and $\beta_1 = 0.9$.

Our $G_{1 \rightarrow 2}$ consists in a stack of 6 convolutional layers with kernel size 3x3 going down to a quarter of the input resolution and then upsampling back to original resolution. We train this network for 100k iterations with batch size 1 and random crops of $512 \times 512$ pixels. We use Adam [1] as optimizer with learning rate $10^{-5}$.

## 3. Details on the evaluation process

We perform all the evaluation at the original image resolution for Cityscapes, Carla and Synthia. Instead, for Kitti, we consider a central crop with size $320 \times 1216$ due to the varying size of images.

**Semantic Segmentation** We train and evaluate the semantic segmentation task on 11 classes, the 10 defined by the Carla framework[2] plus the additional 'Sky' class that we define as the set of points at infinite depth. To evaluate the network on Cityscapes we collapse some of the available classes to make them compatible with Carla: *car* and *bicycle* collapse into *vehicle* and *traffic sign* and *traffic light* into *traffic sign*. We ignore the other labels in Cityscapes which do not have a corresponding class in Carla.

**Depth** We trained and evaluate the depth networks clipping the max predictable depth to 100m and then normalizing between 0 and 1. At inference time we scale the predictions back to the 0m-100m range before computing the different metrics.

---
[2]https://github.com/carla-simulator/carla/releases/tag/0.8.4

| $\mathcal{B} like \mathcal{A}$ inputs | Baseline | CycleGAN | AT/DT | AT/DT+CycleGAN |

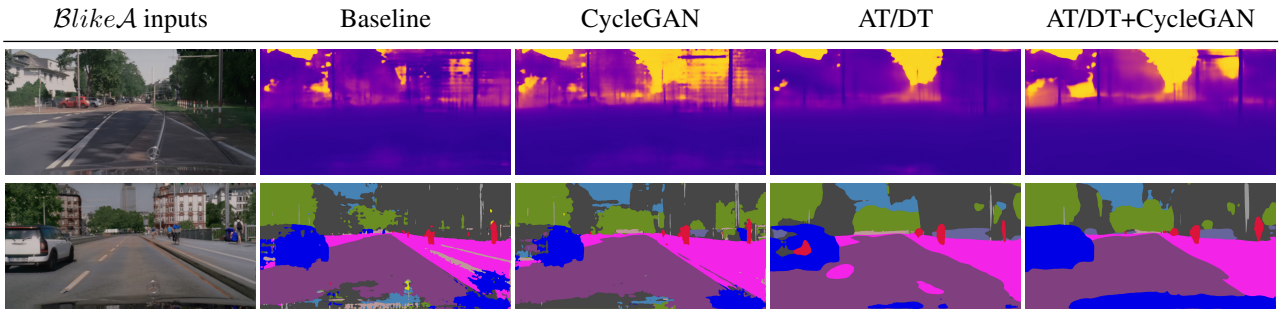

Figure 3: Qualitative results on the Cityscapes dataset in a $Sem. \rightarrow Dep.$ scenario (first row) and $Dep. \rightarrow Sem.$ scenario (second row). From left to right: $\mathcal{B} like \mathcal{A}$ inputs, predictions obtained by a transfer learning baseline, by a domain adaptation baseline (CycleGAN[5]), by our framework (AT/DT) and by our framework aided by domain adaptation (AT/DT+CycleGAN).

| RGB input | Baseline | AT/DT |

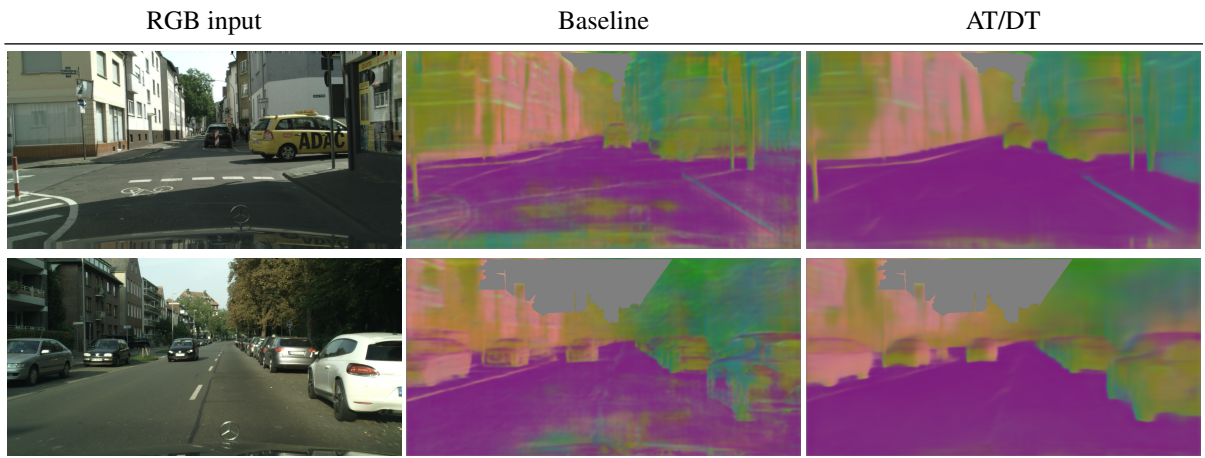

Figure 4: Qualitative results on Cityscapes dataset in a $Sem. \rightarrow Norm.$ scenario. From left to right: RGB input, prediction obtained by a transfer learning baseline and by our framework (AT/DT).


\end{document}